\documentclass{article} 
\usepackage{iclr2027_conference,times}

\usepackage{amsmath,amsfonts,bm}

\def\eqref#1{equation~\ref{#1}}

\def\1{\bm{1}}

\DeclareMathAlphabet{\mathsfit}{\encodingdefault}{\sfdefault}{m}{sl}
\SetMathAlphabet{\mathsfit}{bold}{\encodingdefault}{\sfdefault}{bx}{n}

\usepackage{wrapfig}
\usepackage{hyperref}
\usepackage{url}
\usepackage{booktabs} 
\usepackage{multirow} 
\usepackage{graphicx}
\usepackage{tabularx}
\usepackage{array}
\usepackage{makecell}
\title{EngIntervene: Benchmarking Multimodal Engineering State Understanding and Design Intervention Reasoning}

\author{
Jinchang Zhang, Yingda Tao, Jiakai Lin, Guoyu Lu
\thanks{Corresponding author. Email: \texttt{guoyulu62@gmail.com}}\\
Intelligent Vision and Sensing (IVS) Lab\\
Indiana University Bloomington\\
Bloomington, IN, USA
}

\iclrfinalcopy 
\begin{document}

\maketitle

\begin{abstract}
Multimodal engineering benchmarks largely evaluate static understanding, such as recognizing components, interpreting diagrams, or answering technical questions. This leaves a missing middle between engineering perception and full design generation: whether a model can use an understood system state to reason about relations, constraints, and the consequences of design changes. We introduce \textsc{EngIntervene}, a benchmark for this capability. It contains 3,229 questions across seven engineering domains and organizes evaluation into four levels: state grounding (T1), relational and mechanistic reasoning (T2), constraint-aware diagnosis (T3), and intervention reasoning (T4), which asks whether a proposed modification achieves its target while preserving required constraints. The tasks instantiate a unified engineering state representation spanning objects, relations, constraints, and design objectives, and T2--T4 are scored against structured reference answers with atomic criteria. Across open- and closed-weight multimodal models, stronger grounding does not reliably translate into better diagnosis or intervention, and the best open-weight model trails the best closed model by 14.7 percentage points on the T2--T4 average. Removing or shuffling visual evidence consistently degrades performance, while benchmark-specific supervised fine-tuning improves T1 but not T2--T4. T4 further exposes a large gap between satisfying individual revision criteria and producing a fully valid intervention. Engineering reasoning thus requires not only recovering the current state, but also reliably using it to reason about constraints and post-intervention consequences.
Code and benchmark artifacts are available at
\url{https://github.com/changcv2021/EngIntervene}.
\end{abstract}

\section{Introduction}

\begin{figure*}[h]
\begin{center}
\includegraphics[width=14cm, height=11cm]{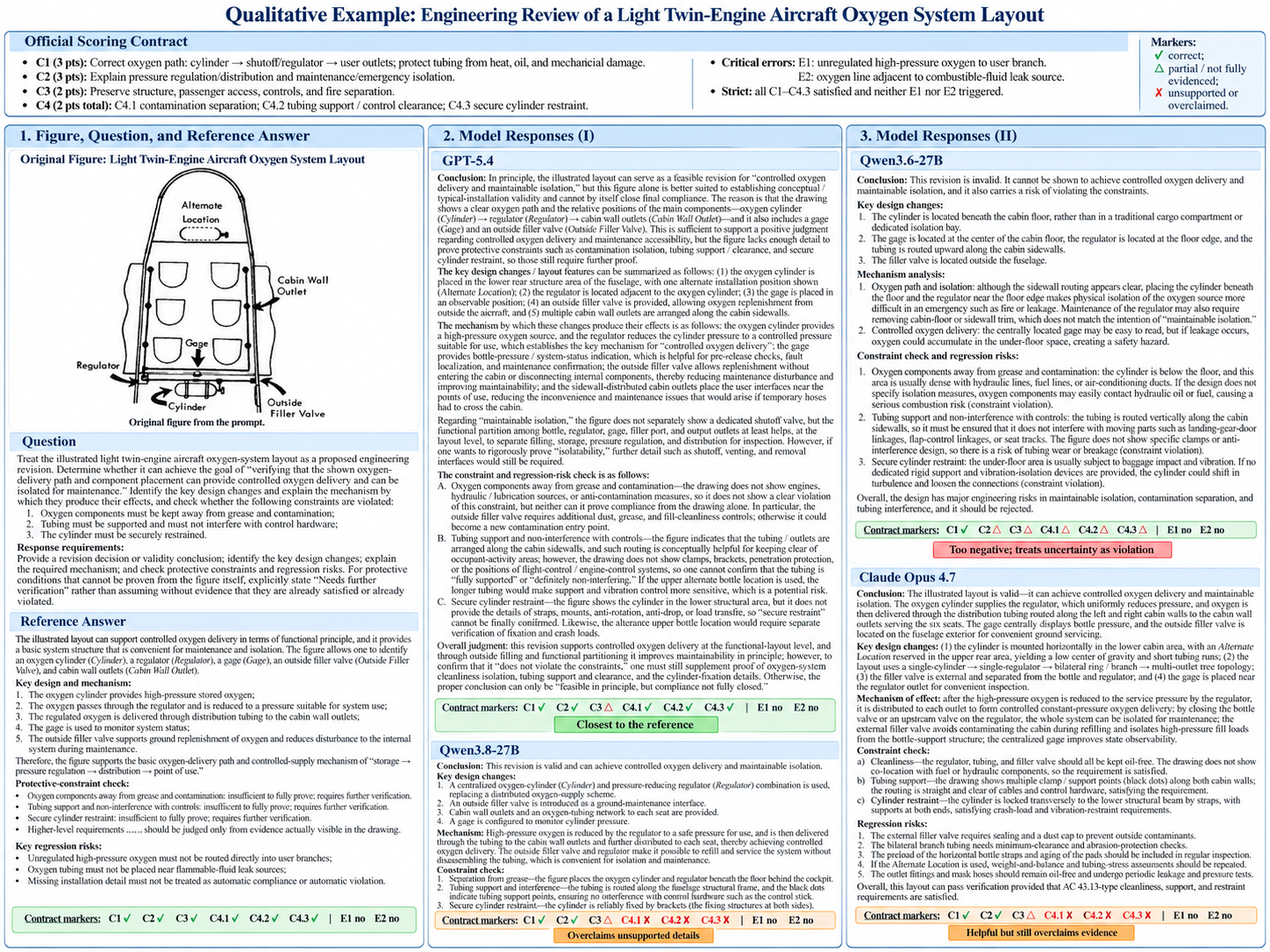}
\end{center}
\vspace{-6mm}
\caption{
Qualitative Example.
}
\vspace{-6mm}
\label{fig:error}
\end{figure*}

Modern engineering systems underpin industrial production, transportation, building operations, energy supply, and electronic information infrastructure \citep{manufacturing, transportation, buildingops, energy, electronicinfoinfra,zhou2025software}. Their design coordinates multiple components, physical processes, and interacting constraints. Although these systems span diverse domains and differ substantially in their physical mechanisms and application contexts, they share common design principles: components, interfaces, and functional modules must work together through appropriate structures, spatial layouts, and transmission paths, while simultaneously satisfying requirements on functional performance, safety and reliability, manufacturability and assembly, maintenance accessibility, and engineering standards. Moreover, local design decisions may alter the functional boundaries and safety state of the entire system. Therefore, understanding engineering design requires more than recognizing ``what is in the system''; it also requires reasoning about ``how the parts interact'' and ``what will happen when the design changes.'' As illustrated in Figure~\ref{fig:error}, across  engineering design scenarios, models may correctly recognize key system components, identify local constraint conflicts, and even propose seemingly reasonable modifications, yet such modifications can still fail in the broader system context. These failures demonstrate that engineering capability cannot be reduced to component recognition, domain-knowledge question answering, or isolated defect detection. Reliable engineering reasoning further requires models to maintain a consistent representation of system entities, relationships, constraints, and design objectives, and to predict how local modifications affect the overall engineering state.

Engineering visual information commonly appears in the form of CAD models, engineering drawings, system schematics, and images of physical equipment. Unlike images, critical engineering information is often encoded in the spatial arrangements among components, connectivity topology, motion relationships, load or energy transmission paths, interface compatibility, and maintenance and safety zones. 
Recovering and using this information is what we call {multimodal engineering state understanding and intervention reasoning}: 
given engineering visual input, a model must recover the current state of the system and determine how a candidate modification changes it.
Existing multimodal and engineering benchmarks cover several adjacent capabilities. MMMU evaluates expert-level knowledge and multimodal problem solving \citep{yue2024mmmumassivemultidisciplinemultimodal}; DesignQA focuses on engineering-rule understanding and design compliance \citep{doris2024designqamultimodalbenchmarkevaluating}; CircuitSense evaluates visual--symbolic reasoning over circuit diagrams \citep{akbari2026circuitsensehierarchicalmllmbenchmark}; and EngDesign assesses open-ended engineering design generation through simulation-based validation \citep{guo2025engineeringagibenchmarkingengineering}.
Between static engineering question answering and complete design synthesis, however, there remains a missing middle: whether models can reason consistently about system relationships, engineering constraints, and the consequences of local design changes.
Evaluating only the final answer or final artifact makes it difficult to determine whether a failure originates from entity recognition, relation modeling, constraint diagnosis, or the prediction of side effects induced by a design change.

To fill this gap, we introduce \textbf{EngIntervene}, a cross-domain benchmark for multimodal engineering state understanding and design-change reasoning. EngIntervene contains {3,229 questions and 3,936 visual inputs} spanning seven engineering domains: mechanical design and industrial equipment, robotics and mechatronic systems, automotive system integration, aerospace system integration, building mechanical, electrical, and plumbing (MEP) systems, energy and power equipment, and compact electronic packaging. The benchmark decomposes the target capability into four progressively demanding tasks:
\text{T1 Understanding},
\text{T2 Relation Modeling}
,
\text{T3 Diagnosis}
,
\text{T4 Change Reasoning}.
T1 evaluates the recognition of engineering entities, functions, and design intent; T2 evaluates reasoning about connections, paths, motion, and operating mechanisms; T3 evaluates problem localization, constraint identification, and risk analysis; and T4 evaluates whether a candidate modification achieves its intended objective while preserving critical constraints and avoiding newly introduced problems. T1 uses multimodal multiple-choice questions, whereas T2--T4 use open-ended responses.
To ensure traceability, the questions, reference answers, and scoring criteria in EngIntervene are grounded in university examinations, textbooks, professional qualification examinations, technical training materials, design guidelines, and engineering case studies. For T2--T4, we construct and freeze question-level atomic scoring rubrics before evaluation, decomposing each reference answer into independent scoring items covering required entities, relationships, constraints, underlying mechanisms, and critical errors. Open-ended responses are evaluated using a unified automatic grading protocol that determines whether each scoring criterion is explicitly satisfied and requires every positive judgment to be supported by verbatim evidence from the evaluated response.
EngIntervene thus shifts multimodal engineering evaluation from asking ``can the model answer questions about the current design?'' to asking ``can the model determine what will happen when the design changes?''

Our contributions are as follows:
1. We introduce {EngIntervene}, a multimodal engineering reasoning benchmark spanning seven engineering domains.
    2.  We formalize engineering design reasoning into four progressively structured capability levels under a unified engineering-state representation, enabling separate evaluation of a model's ability to understand the current state of an engineering system and to use that state for relational reasoning, constraint-aware diagnosis, and design intervention.
    3. Through experiments, we reveal a gap between the ``understanding'' and ``use'' of engineering states. Stronger state understanding does not reliably translate into better diagnosis and intervention reasoning, and improvements in basic understanding from domain-specific supervised fine-tuning do not automatically transfer to reasoning gains on T2--T4.

\begin{figure}[t]
    \centering
     \includegraphics[width=14cm, height=8cm]{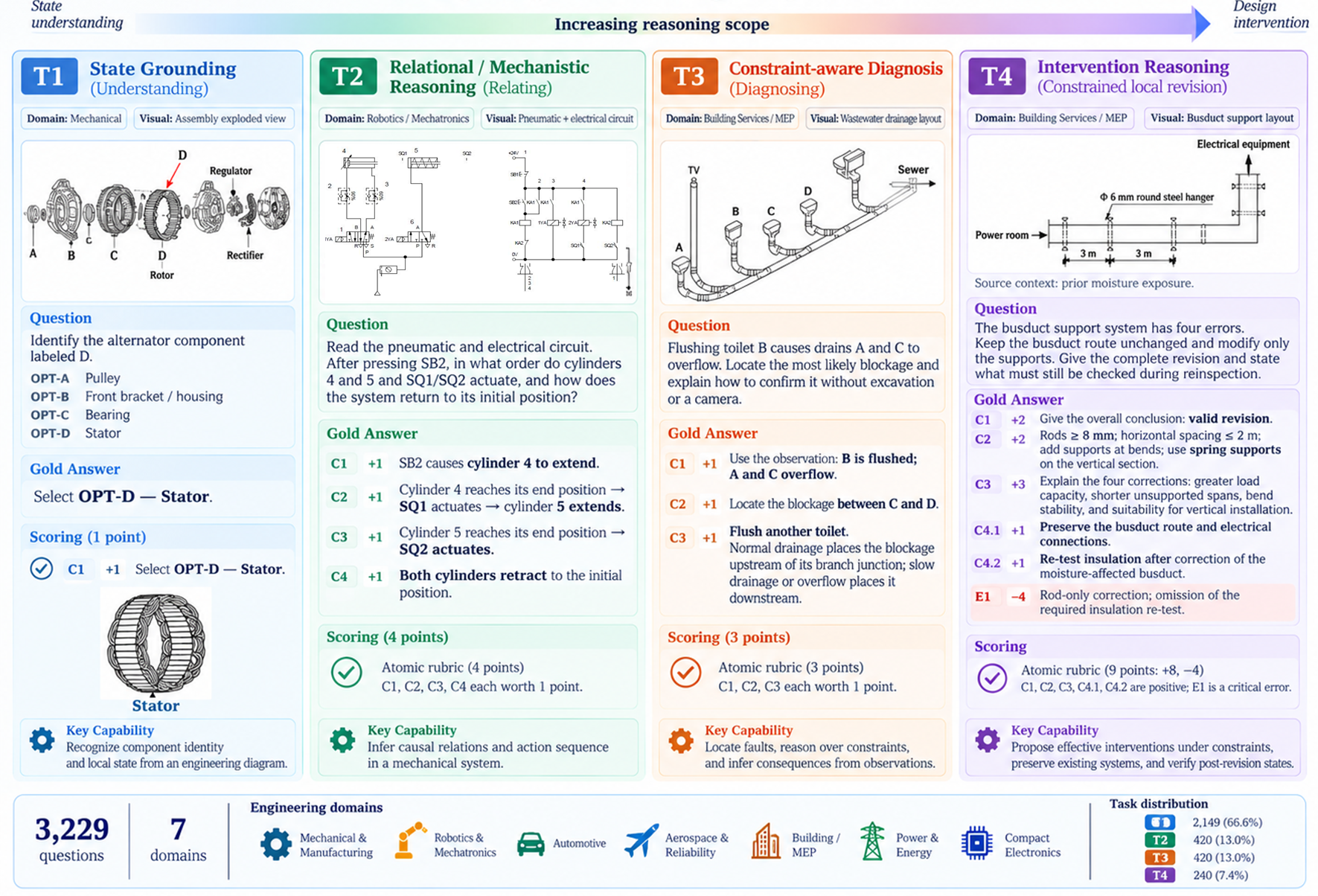}
     \vspace{-8mm}
    \caption{\textbf{Overview of EngIntervene.}
    The center illustrates the four T1--T4 capability levels with representative real examples, while the outer ring presents the seven engineering domains.}
    \label{fig:engintervene_overview}
      \vspace{-8mm}
\end{figure}

\vspace{-3mm}
\section{EngIntervene}
\label{sec:benchmark}
\vspace{-1mm}
\subsection{Benchmark Overview and Unified Problem Formulation}
\label{sec:benchmark_overview}

We introduce \textbf{EngIntervene}, a benchmark for evaluating engineering state understanding and design intervention reasoning in multimodal large language models. Unlike existing multimodal engineering tasks that primarily focus on object recognition, diagram-based question answering, or domain-knowledge recall, EngIntervene addresses a more general question: whether a model can recover the current state of an engineering system from visual and textual evidence, identify its structural relations and validity constraints, and further reason about how local design interventions affect the system state.
Engineering decision making typically goes beyond identifying a component or explaining the function of a mechanism. It requires simultaneously understanding system composition, dependencies among components, applicable constraints, and design objectives, while determining whether a local modification truly resolves the target problem, preserves existing constraints, and avoids introducing new failure modes elsewhere in the system. EngIntervene therefore extends multimodal engineering reasoning from static ``recognition and explanation'' toward design-oriented ``diagnosis and intervention.''
To capture the common task structure across engineering domains, we conceptually represent the engineering state involved in a question as
$
\mathcal{S}=(\mathcal{V},\mathcal{E},\mathcal{C},\mathcal{O}),
$
where \(\mathcal{V}\) denotes relevant objects and their functional attributes, \(\mathcal{E}\) denotes structural and process relations among objects, \(\mathcal{C}\) denotes engineering constraints applicable to the current problem, and \(\mathcal{O}\) denotes the objectives or priority conditions specified by the question. A design intervention \(\Delta\) maps the current state to a modified state,
$
\mathcal{S}'=\tau(\mathcal{S},\Delta).
$
Under this formulation, a valid design intervention must satisfy two conditions simultaneously: the modified state must achieve the specified objective, and the constraints that are required to remain satisfied must still hold after the modification,
$
\mathrm{Valid}(\Delta)
=
\mathrm{Goal}(\mathcal{S},\mathcal{S}')
\;\land\;
\bigwedge_{c\in\mathcal{C}_p} c(\mathcal{S}'),
$
where \(\mathcal{C}_p\subseteq\mathcal{C}\) denotes the constraints that must continue to hold after the modification, including explicitly stated protected conditions and hard constraints applicable to the system. New failure modes or regressions introduced elsewhere in the system correspond to cases in which some \(c\in\mathcal{C}_p\) no longer holds in \(\mathcal{S}'\); they therefore fall under the second condition rather than forming a separate third condition. Design intervention reasoning, evaluated in T4 below, asks whether the model can determine whether \(\mathrm{Valid}(\Delta)\) holds by both predicting the state change induced by \(\Delta\) and examining its effects on \(\mathcal{C}_p\).

Based on this structure, we divide the evaluation into four progressively structured capability levels. \textbf{T1 (State Grounding)} evaluates whether the model can identify engineering objects, component functions, and local design states. \textbf{T2 (Relational and Mechanistic Reasoning)} evaluates whether the model can recover connectivity, spatial, motion, and process relations among objects and explain how the system operates. \textbf{T3 (Constraint-aware Diagnosis)} requires the model to identify failures or invalid states in the current design and connect them to specific engineering constraints and potential consequences. \textbf{T4 (Intervention Reasoning)} further requires the model to reason about the state transition induced by a design modification, determine whether the target problem is resolved, assess whether critical constraints remain satisfied, and identify whether the modification introduces new regressions or design conflicts.
This organization describes different levels of engineering reasoning capability; it does not require each individual sample to be solved sequentially from T1 through T4. The seven engineering domains are instead treated as domain-specific instantiations of this unified capability structure under different systems, objects, and constraints.

\vspace{-3mm}
\subsection{Four Tasks and Cross-domain Instantiation}
\vspace{-3mm}
\label{sec:benchmark_tasks}
\begin{wraptable}{r}{0.52\textwidth}
\vspace{-8mm}
\centering
\caption{{The four EngIntervene tasks and their core reasoning objectives.}}
\label{tab:task_overview}

\setlength{\tabcolsep}{3pt}
\renewcommand{\arraystretch}{1.10}

\resizebox{0.5\textwidth}{!}{
\begin{tabular}{
@{}
>{\raggedright\arraybackslash}m{0.25\textwidth}
>{\centering\arraybackslash}m{0.2\textwidth}
>{\raggedright\arraybackslash}m{0.29\textwidth}
>{\raggedright\arraybackslash}m{0.35\textwidth}
@{}
}
\toprule

\textbf{Task}
&
\makecell[c]{\textbf{Engineering State}\\\textbf{Components}}
&
\makecell[c]{\textbf{Primary Response}\\\textbf{Format}}
&
\makecell[c]{\textbf{Primary Evaluation}\\\textbf{Target}}
\\

\midrule

T1: State Grounding
& \(\mathcal{V}\)
& Multiple choice
& Objects, functions, states, and source-supported design intent
\\

\makecell[l]{T2: Relational /\\Mechanistic Reasoning}
& \(\mathcal{V},\mathcal{E}\)
& \makecell[l]{Natural-language\\explanation}
& \makecell[l]{Connections, paths, sequences, and\\operating mechanisms}
\\

\makecell[l]{T3: Constraint-aware\\Diagnosis}
& \(\mathcal{V},\mathcal{E},\mathcal{C}\)
& \makecell[l]{Natural-language\\diagnosis Revision}
& \makecell[l]{Problem location, violated conditions,\\and engineering consequences}
\\

\makecell[l]{T4: Intervention\\Reasoning}
& \(\mathcal{V},\mathcal{E},\mathcal{C},\mathcal{O}\)
& \makecell[l]{ selection, verification, local\\proposal, or prioritization\\ withjustification}
& \makecell[l]{Target effects, protected constraints,\\and unacceptable side effects}
\\

\bottomrule
\end{tabular}
}
\vspace{-4mm}
\end{wraptable}
The four tasks share the engineering-state representation introduced in Sec.~\ref{sec:benchmark_overview}, but focus on different state components and reasoning operations. Table~\ref{tab:task_overview} summarizes the correspondence between each task and the unified state representation.
\textbf{T1: Multimodal Design Understanding.}
T1 focuses on recognizing and grounding engineering objects and their functions, including component categories, module purposes, local structures, and design intent explicitly supported by the source material.
\textbf{T2: Mechanistic and Relational Reasoning.}
T2 extends object recognition to understanding internal system relations. Models must reason about connections, transmission and operational relations among objects, including relevant entities, structural relations, paths, directions, and process sequences. T2 focuses on how a system is organized and operates, rather than whether the current design is appropriate.
\textbf{T3: Constraint-aware Design Diagnosis.}
T3 extends system understanding to validity assessment. Models must identify problems in the current design, locate the relevant objects or regions, determine the violated engineering constraints or design principles, and explain the resulting functional, safety, manufacturing, reliability, or maintenance consequences.
\textbf{T4: Constraint-preserving Design Intervention Reasoning.}
T4 contains four complementary intervention-reasoning operations.

\textbf{Revision Selection.}
Given two or more candidate modifications, the model compares the potential post-intervention states and selects the option that best satisfies the target and constraints.
\textbf{Revision Verification and Regression Analysis.}
Given a specific modification, the model determines whether it fully resolves the target problem while identifying whether it violates other existing conditions or introduces new failure modes.
\textbf{Constrained Local Revision Proposal.}
Given a clearly specified problem, an editable local region, and acceptance conditions, the model proposes a local design modification that achieves the target without violating protected constraints.

\textbf{Revision Prioritization and Trade-off.}
When multiple problems or modification directions coexist, the model determines an appropriate intervention order or design trade-off according to hard constraints, risks, and target priorities.
These four operations correspond to different treatments of \(\Delta\). Revision Selection chooses a \(\Delta\) satisfying \(\mathrm{Valid}\) from a given 
\begin{wraptable}{r}{0.54\textwidth}
\vspace{-7mm}
\centering
\caption{\textbf{T4 design-intervention reasoning types.}}
\label{tab:t4_taxonomy}
\resizebox{\linewidth}{!}{
\begin{tabular}{lrrr}
\toprule
\textbf{T4 Type} &
\textbf{\# Items} &
\textbf{\% of T4} &
\textbf{\# Domains} \\
\midrule
Revision Selection
& 61 & 25.4\% & 5 \\
Revision Verification and Regression Analysis
& 86 & 35.8\% & 6 \\
Constrained Local Revision Proposal
& 68 & 28.3\% & 6 \\
Revision Prioritization and Trade-off
& 25 & 10.4\% & 4 \\
\bottomrule
\end{tabular}
}
\vspace{-6mm}
\end{wraptable}
candidate set; Revision Verification and Regression Analysis determines whether a given \(\Delta\) satisfies \(\mathrm{Valid}\) and identifies any violated \(c\in\mathcal{C}_p\); Constrained Local Revision Proposal constructs a \(\Delta\) satisfying \(\mathrm{Valid}\) within the allowed editable scope; and Revision Prioritization and Trade-off ranks multiple problems or candidate interventions according to hard constraints and the priorities encoded in \(\mathcal{O}\).
Table~\ref{tab:t4_taxonomy} summarizes the composition of the 240 T4 questions. The full domain-by-T4-subtask distribution is reported in Appendix~\ref{app:t4_distribution}.

\vspace{-3mm}
\subsection{Data Collection, Answer Representation, and Quality Control}
\vspace{-3mm}
\label{sec:data_construction}

Candidate materials for EngIntervene are collected from examinations and qualification materials, university courses and textbooks, technical manuals, manufacturer design guidelines, and case materials with explicit engineering conclusions, covering 357 source documents. Questions are constructed in two ways: normalization of existing source questions, and constrained question construction and adaptation based on reliable source materials. Among the 3,229 questions, 2,850 are obtained by normalizing existing source questions and 379 are source-supported constructed questions; definitions and task-level statistics for the two construction types are provided in Appendix~\ref{app:construction_stats}.
The construction pipeline includes source discovery, candidate-question extraction, visual-material matching, task assignment, reference-answer preparation, structured scoring-target construction, automated structural checks, and human semantic review. For constructed questions, newly introduced content is restricted to facts, constraints, and engineering conclusions supported by the source material, avoiding unsupported inference.

\textbf{Structured Reference Answers: }
Different tasks use reference-answer representations aligned with their reasoning objectives. T1 stores stable option identifiers and necessary explanations so that semantic correctness remains unchanged when option positions vary. T2 represents the entities, critical relations, paths or process sequences required for a complete explanation, together with critical physical errors that should be excluded. T3 decomposes design diagnosis into problem location, problem type, corresponding engineering constraint, and potential engineering consequence. T4 organizes its reference representation around the target problem, modification, mechanism, conditions that must remain satisfied, acceptable equivalent expressions, and unacceptable regressions. These structures support subsequent atomic evaluation, allowing natural-language answers to be scored according to engineering semantics rather than surface-form similarity.

\textbf{Atomic Scoring Targets: }
For T2--T4, we further convert the reference answers into atomic scoring criteria aligned with the task requirements. T4 evaluates not only whether the target modification is correct, but also explicitly records the modification objective, engineering conditions that must remain satisfied, and critical errors or regressions that may invalidate the modification. For compound protected conditions that can be directly aligned with the original requirements, we decompose them into atomic leaf criteria: 118 T4 questions contain such compound conditions, yielding 279 leaf criteria in total. The weights of the leaf criteria sum to the weight of the original compound criterion. This decomposition enables revision validity and constraint preservation to be evaluated separately.

\begin{wraptable}{r}{0.49\textwidth}
\vspace{-6mm}
\centering
\caption{{Independent quality audit of EngIntervene before revision.}}
\label{tab:quality_audit}

\setlength{\tabcolsep}{3pt}        
\renewcommand{\arraystretch}{0.95} 

\resizebox{\linewidth}{!}{
\begin{tabular}{lrrrr}
\toprule
\textbf{Audit Dimension}
& \textbf{\# Audited}
& \textbf{\# Passed}
& \textbf{Pass Rate}
& \textbf{95\% CI} \\
\midrule
Source support
& 500 & 493 & 98.6\% & 97.1--99.3 \\
Image--question consistency
& 500 & 489 & 97.8\% & 96.1--98.8 \\
Reference-answer correctness
& 500 & 490 & 98.0\% & 96.4--98.9 \\
Question unambiguity / answerability
& 500 & 481 & 96.2\% & 94.1--97.6 \\
Rubric completeness (T2--T4)
& 300 & 285 & 95.0\% & 91.9--96.9 \\
All applicable criteria satisfied
& 500 & 474 & 94.8\% & 92.5--96.4 \\
\bottomrule
\end{tabular}
}

\vspace{-6mm}
\end{wraptable}
\textbf{Quality Control:} We stratify 500 questions from the dataset by task type and construction method and have them reviewed item by item by seven reviewers whose engineering backgrounds match the question's domain and who did not construct the corresponding samples. Each question is checked on five dimensions: source support, image--question consistency, reference-answer correctness, question answerability, and T2--T4 rubric completeness (Table~\ref{tab:quality_audit}). Before revision, 94.8\% of audited questions satisfy all applicable criteria, and the lowest per-dimension pass rate is 95.0\% (rubric completeness). Failures mainly involve local wording ambiguity, incomplete rubric coverage, or missing protected conditions; all are revised and re-reviewed. All 240 T4 questions additionally undergo a dedicated design-intervention audit (Appendix~\ref{app:t4_audit}).

\vspace{-3mm}
\subsection{Coverage and Diversity of the Engineering Reasoning Space}
\label{sec:coverage_diversity}
\vspace{-1mm}
We characterize benchmark coverage along five dimensions: engineering domains, task capabilities, shared engineering constraints, visual representations, and source structure.

\begin{figure}[t]
    \centering
    \begin{minipage}[t]{0.49\linewidth}
        \centering
        \includegraphics[width=\linewidth]{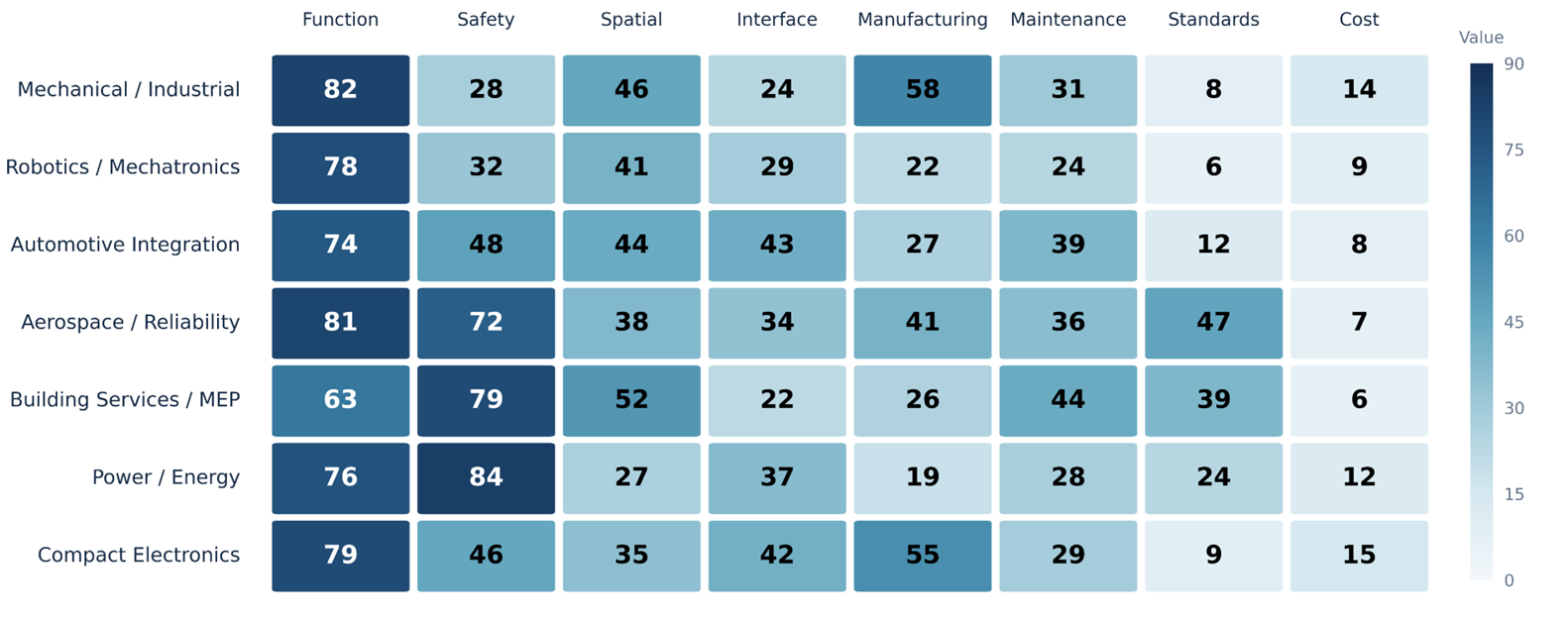}
        \vspace{-7mm}
        \caption{{Shared engineering constraint coverage across domains.}
        }
        \label{fig:constraint_heatmap}
        \vspace{-7mm}
    \end{minipage}
    \hfill
    \begin{minipage}[t]{0.49\linewidth}
        \centering
        \includegraphics[width=\linewidth]{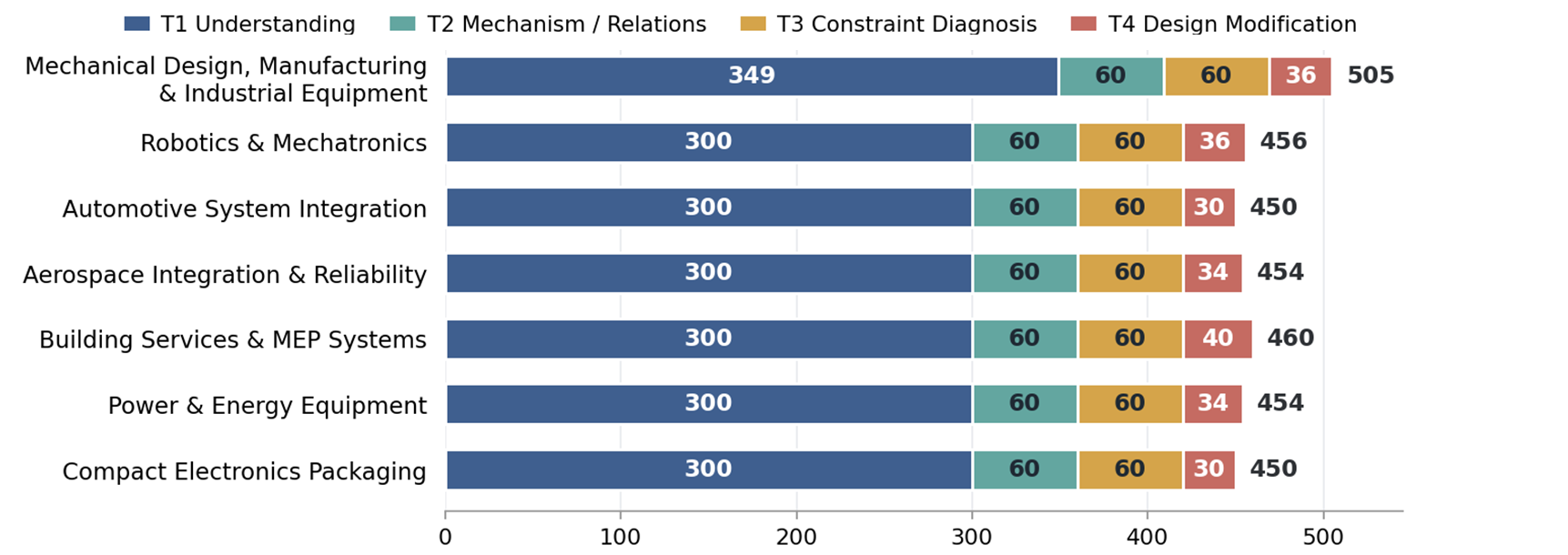}
        \vspace{-7mm}
        \caption{{Domain-by-task distribution of EngIntervene.}
        }
   \label{tab:domain_task_distribution}
   \vspace{-6mm}
    \end{minipage}
\end{figure}

\textbf{Unified Capability Instantiation across Domains.}
EngIntervene contains 3,229 questions across seven engineering domains. Fig.~\ref{tab:domain_task_distribution} reports the domain-by-task distribution. Each domain contains between 450 and 505 questions, preventing any single engineering domain from dominating the overall evaluation. T2 and T3 maintain balanced coverage across all seven domains, while T4 includes multiple forms of intervention reasoning, including revision selection, revision verification, constrained local revision, and design trade-offs. The full domain-by-T4-reasoning-type distribution is provided in Appendix~\ref{app:t4_distribution}.
Figure~\ref{fig:constraint_heatmap} further illustrates the coverage of shared engineering constraints across the seven domains. Although the domains involve different objects, system forms, and visual representations, Function / Performance constraints cover 63.0\%--87.8\% of T3--T4 questions in every domain, while Spatial / Clearance, Interface / Compatibility, Manufacturability / Assembly, and Maintainability / Serviceability each cover 20.2\%--61.1\% across all domains. Safety / Reliability constraints are most concentrated in Aerospace, Building Services, and Power and Energy (76.6\%--89.4\%), and Standards / Regulation constraints also occur primarily in these more regulation-intensive domains. Cost / Resource constraints account for no more than 16.7\% in any domain. On average, each T3--T4 question involves 3.2 constraint categories. The seven domains therefore do not form seven independent question sets, but rather instantiate a shared set of relations, constraints, and design-intervention structures in different engineering systems. Full statistics are provided in Appendix~\ref{app:constraint_coverage}.


\textbf{Heterogeneous Visual Representations.}
EngIntervene contains 2,275 distinct images used by benchmark questions, covering CAD views, engineering drawings, assembly drawings, mechanism diagrams, system schematics, piping and wiring diagrams, BIM/MEP layouts, before/after design images, and real-equipment images (Appendix~\ref{app:visual_representation}).
Because the same engineering image may support questions at different reasoning levels, an image can be referenced by multiple questions. Across the benchmark, the questions contain 3,936 image references in total, and 483 questions (15.0\%) require reasoning jointly over multiple images. Models must therefore handle not only single-image object recognition, but also cross-view correspondence, state comparison, and multi-image relational reasoning. Task-level visual-input composition and question-length statistics are provided in Appendix~\ref{app:input_statistics}.

\textbf{Source Structure and Content Diversity.}
The 357 source documents associated with EngIntervene form 227 related source groups after accounting for relationships among materials. The largest individual source contributes 7.7\% of all questions. The dataset therefore covers multiple textbook systems, engineering guidelines, technical manuals, certification materials, and case materials. Complete statistics on source types, source languages, construction methods, and source concentration are provided in Appendices~\ref{app:construction_stats}--\ref{app:source_language} and Appendix~\ref{app:source_concentration}.

\textbf{T4 Composition and Construction.}
The detailed distribution of the four T4 design-intervention task types across the seven engineering domains is provided in Appendix~\ref{app:t4_distribution}.
Among the 240 T4 questions, 132 (55.0\%) are source-supported constructed items (Appendix~\ref{app:construction_stats}).
These questions are constructed from traceable engineering materials under explicit modification objectives and acceptance conditions; their reference modifications, protected constraints, and regression conditions are examined through a dedicated T4 audit (Appendix~\ref{app:t4_audit}).


\vspace{-2mm}
\section{Experiment}

\subsection{Main Results} 
\vspace{-1mm}
Table~\ref{tab:main_results} summarizes the performance of closed and
open-weight models on EngIntervene.
T1 accuracy is substantially above the four-choice random baseline for
all models, and T1 is generally positively correlated with the T2--T4
average across models. Models with similar T1 scores can nevertheless
differ substantially on the open-ended tasks: Qwen3-VL-8B-Instruct and
Qwen3-VL-8B-Thinking differ by 0.47 points on T1 but by 15.19 points on
T4 Atomic, in the opposite direction. Object- and function-level
grounding is therefore necessary but not sufficient for open-ended
engineering reasoning.
Claude Opus 4.7 achieves 57.64\%, 58.41\%, and 67.84\% on T2, T3, and
T4 Atomic, respectively, with a T2--T4 average of 61.30\%, the highest
among all evaluated models. The strongest open-weight model,
Qwen3.8-27B, achieves 46.64\% on the same average, leaving a gap of
14.66 percentage points.
T4 Atomic is higher than T2 and T3 for all but one model, partly because
T4 rubrics include goal and protection criteria that are largely
recoverable from the question text (Section~\ref{sec:visual_ablation}).
Strict Success removes this partial credit: Claude Opus 4.7 drops from
67.84\% T4 Atomic to 30.83\% Strict Success, and GPT-5.4 from 62.73\%
to 24.17\%. Section~\ref{sec:t4_diagnostic} analyzes this gap between
partial correctness and complete revision success.

\textbf{Scaling Effects within Model Families. }
Within the Qwen3.5 family, performance on all four tasks increases from
4B to 9B and 27B: the four-task macro average rises from 38.27\% to
40.90\% and 48.44\%, T4 Atomic from 38.33\% to 40.60\% and 52.12\%, and
T4 Strict from 8.33\% to 11.25\% and 17.92\%. InternVL3.5 also improves
overall from 8B to 14B.

\textbf{Capability Trade-offs across Model Generations.}
At the same nominal 27B scale, Qwen3.5, Qwen3.6, and Qwen3.8 improve on
different tasks. Qwen3.6-27B achieves the highest T1 score among the
three, at 61.66\%, but performs below Qwen3.5-27B on T2--T4.
Qwen3.8-27B obtains a lower T1 score of 58.07\% but reaches 41.03\%,
44.90\%, and 53.98\% on T2, T3, and T4 Atomic, respectively, all higher
than Qwen3.5-27B.

\begin{table*}[t]
\centering
\caption{
Main results on EngIntervene.
T4 Strict denotes the percentage of T4 examples satisfying all required
conditions without triggering a critical error.
}
\label{tab:main_results}
\resizebox{0.9\textwidth}{!}{
\begin{tabular}{llrrrrrrr}
\toprule
\textbf{Type}
& \textbf{Model}
& \textbf{T1 Acc.}
& \textbf{T2 Atomic}
& \textbf{T3 Atomic}
& \textbf{T4 Atomic}
& \textbf{T4 Strict}
& \textbf{Avg. T2--T4}
& \textbf{Avg. T1--T4} \\
\midrule

\multirow{5}{*}{\textbf{Closed}}
& GPT-4o-mini
& 47.98
& 23.96
& 23.01
& 31.29
& 7.50
& 26.09
& 31.56 \\

& GPT-5.4
& 72.68
& 54.40
& 52.33
& 62.73
& 24.17
& 56.48
& 60.53 \\

& GPT-5.6 Luna
& 66.68
& 51.73
& 50.15
& 59.16
& 24.17
& 53.68
& 56.93 \\

& Claude Opus 4.7
& 73.00
& 57.64
& 58.41
& 67.84
& 30.83
& 61.30
& 64.22 \\

& Grok 4.3 (non-reasoning)
& 53.90
& 34.05
& 33.96
& 40.67
& 12.08
& 36.23
& 40.65 \\

\midrule

\multirow{12}{*}{\textbf{Open}}
& Qwen3.5-4B
& 51.65
& 29.78
& 33.32
& 38.33
& 8.33
& 33.81
& 38.27 \\

& Qwen3.5-9B
& 55.37
& 33.10
& 34.54
& 40.60
& 11.25
& 36.08
& 40.90 \\

& Qwen3.5-27B
& 60.31
& 40.40
& 40.93
& 52.12
& 17.92
& 44.48
& 48.44 \\

& Qwen3.6-27B (non-thinking)
& 61.66
& 36.85
& 37.81
& 49.04
& 13.33
& 41.23
& 46.34 \\

& Qwen3.8-27B (non-thinking)
& 58.07
& 41.03
& 44.90
& 53.98
& 18.33
& 46.64
& 49.50 \\

& Qwen3-VL-8B-Instruct
& 49.70
& 30.85
& 28.61
& 27.51
& 6.25
& 28.99
& 34.16 \\

& Qwen3-VL-2B-Thinking
& 46.11
& 18.58
& 21.85
& 29.63
& 6.25
& 23.35
& 29.04 \\

& Qwen3-VL-8B-Thinking
& 49.23
& 30.25
& 32.64
& 42.70
& 11.25
& 35.20
& 38.71 \\

& InternVL3.5-8B
& 45.00
& 22.33
& 24.02
& 31.84
& 4.58
& 26.06
& 30.80 \\

& InternVL3.5-14B
& 47.23
& 23.84
& 25.29
& 32.99
& 5.00
& 27.37
& 32.34 \\

& Gemma 4-31B-IT
& 58.77
& 41.17
& 40.89
& 51.32
& 16.67
& 44.46
& 48.04 \\

& MiMo-VL-7B-RL
& 49.98
& 29.85
& 33.21
& 45.83
& 11.67
& 36.30
& 39.72 \\

\bottomrule
\end{tabular}
}

\vspace{-9mm}
\end{table*}

\vspace{-3mm}

\subsection{Is Visual Evidence Necessary?}
\label{sec:visual_ablation}
\vspace{-2mm}
Multimodal engineering benchmark should require models to use visual evidence corresponding to the current question, rather than relying primarily on textual priors. We therefore evaluate the extent to which models actually depend on visual information. \textbf{Full} uses the text question together with its corresponding images. \textbf{Text-only} preserves all textual information while removing all visual inputs. \textbf{Shuffled-image} preserves the text but replaces the original images with images from other samples in the same task type and engineering domain. This condition maintains visual plausibility at the domain and task levels while breaking the correspondence between the images and the question.








\textbf{Correct visual evidence consistently improves model performance.}

\begin{wraptable}{r}{0.52\textwidth}
\vspace{-5mm}
\centering
\caption{Visual-dependency  on EngIntervene.}
\label{tab:visual_ablation_full}

\resizebox{0.5\textwidth}{!}{
\begin{tabular}{llrrrrrr}
\toprule
Model
& Condition
& T1
& T2
& T3
& T4 Atomic
& T4 Strict
& Macro Avg. \\
\midrule

\multirow{3}{*}{Qwen3.5-27B}
& Full
& 60.03 & 40.40 & 40.93 & 52.12 & 17.92 & 48.37 \\
& Text-only
& 39.37 & 29.70 & 34.59 & 46.87 & 15.00 & 37.63 \\
& Shuffled-image
& 31.74 & 7.02 & 16.11 & 26.75 & 6.67 & 20.40 \\
\midrule

\multirow{3}{*}{Qwen3.6-27B}
& Full
& 61.66 & 36.85 & 37.81 & 49.04 & 13.33 & 46.34 \\
& Text-only
& 37.65 & 25.63 & 33.06 & 46.49 & 10.42 & 35.71 \\
& Shuffled-image
& 36.25 & 8.70 & 17.62 & 27.04 & 7.50 & 22.40 \\
\midrule

\multirow{3}{*}{Qwen3.8-27B}
& Full
& 58.07 & 41.03 & 44.90 & 53.98 & 18.33 & 49.50 \\
& Text-only
& 37.97 & 26.24 & 31.49 & 48.99 & 13.75 & 36.17 \\
& Shuffled-image
& 36.20 & 15.67 & 25.38 & 39.85 & 11.67 & 29.27 \\
\midrule

\multirow{3}{*}{Gemma 4-31B-IT}
& Full
& 58.63 & 41.17 & 40.89 & 51.32 & 16.67 & 48.00 \\
& Text-only
& 38.99 & 29.67 & 34.28 & 51.27 & 16.25 & 38.55 \\
& Shuffled-image
& 30.39 & 6.18 & 10.62 & 26.87 & 5.00 & 18.52 \\
\midrule

\multirow{3}{*}{MiMo-VL-7B-RL}
& Full
& 49.79 & 29.30 & 32.26 & 45.55 & 11.67 & 39.22 \\
& Text-only
& 34.53 & 22.40 & 26.56 & 42.04 & 12.08 & 31.38 \\
& Shuffled-image
& 35.18 & 15.77 & 20.22 & 39.52 & 8.75 & 27.67 \\

\bottomrule
\end{tabular}
}

\vspace{-6pt}
\end{wraptable}
Removing the correct images lowers the four-task macro average of all
five models in Table~\ref{tab:visual_ablation_full}, by 10.74
(Qwen3.5-27B), 10.63 (Qwen3.6-27B), 13.33 (Qwen3.8-27B), 9.45
(Gemma 4-31B-IT), and 7.84 (MiMo-VL-7B-RL) percentage points.
Under Shuffled-image, performance drops further. Taking the difference
between Text-only and Shuffled-image as the interference caused by
incorrect images, the five models lose a further 17.23, 13.31, 6.90,
20.03, and 3.71 percentage points, respectively. The interference a
model suffers is not predicted by its gain from correct images.
Gemma 4-31B-IT gains less than Qwen3.8-27B from correct images
(9.45 vs.\ 13.33 points) but suffers nearly three times the interference
(20.03 vs.\ 6.90 points); Qwen3.5-27B and Qwen3.8-27B, both nominally
27B, differ by more than 10 points in interference (17.23 vs.\ 6.90).
The interference is most pronounced on T2: the five-model mean T2 score
is 26.73\% under Text-only but falls to 10.67\% under Shuffled-image.
Models describe components and relations from the incorrect images that
do not exist in the corresponding questions, thereby triggering critical
error criteria.

\textbf{Different tasks exhibit stable but substantially different patterns of visual dependence.}

From Full to Text-only, all five models exhibit a consistent ordering in performance degradation: T1 $>$ T2 $>$ T3 $>$ T4 Atomic. T1 decreases by 15.26--24.01 percentage points, substantially more than T2 (6.90--14.79), T3 (4.75--13.41), and T4 Atomic (0.05--5.25). This pattern is consistent with the task structure: T1 directly depends on objects, functions, and states in engineering figures, whereas T2 and T3 jointly rely on visual structure and engineering conditions specified in the text.
T4 exhibits a different pattern of visual dependence. After removing the image, the decrease in Atomic Score is relatively small. For example, Qwen3.5-27B drops from 52.12\% to 46.87\%, while Gemma 4-31B-IT remains nearly unchanged (51.32\%$\rightarrow$51.27\%). This suggests that explicitly stated design objectives, candidate modifications, and protected constraints in the text can support partial judgments about a revision. However, such local credit masks a larger degradation in complete intervention validity: the average Strict Success across the five models decreases from 15.58\% to 13.50\%, corresponding to a relative drop of 13.4\%. Thus, although textual information can partially compensate for missing visual evidence in local T4 judgments, verifying a fully valid design intervention still depends on visual evidence.
The Shuffled-image condition further amplifies this gap. Averaged across the five models, T4 Atomic decreases from 50.40\% to 32.01\%, while Strict Success drops from 15.58\% to 7.92\%, a relative decrease of 49.2\%. This indicates that inconsistent visual evidence not only weakens local engineering judgments, but more severely disrupts the ability to satisfy the complete set of intervention requirements simultaneously. 
\vspace{-3mm}
\subsection{Fine-grained Analysis of Design-Change Reasoning}
\label{sec:t4_diagnostic}
\vspace{-2mm}
The overall T4 score measures performance on design-change reasoning,
but does not reveal why a revision judgment succeeds or fails.
We therefore conduct a post-hoc diagnostic analysis using the existing
criterion-level judgments and group independently evaluable rubric
criteria into four functional roles:
{Goal Achievement},
{Mechanism Reasoning},
{Constraint Preservation},
and {Regression Avoidance}.
Goal measures whether a model correctly selects, proposes, rejects, or
qualifies a modification and judges its intended effect;
Mechanism measures whether it explains why the modification produces
that effect;
Protection measures whether required non-target conditions remain
satisfied;
and Regression captures explicit checking of residual or newly
introduced risks.
Complete role-wise results are reported in
Appendix~\ref{app:t4_role}.

\begin{figure}[t]
    \centering
    \begin{minipage}[t]{0.49\linewidth}
        \centering
        \includegraphics[width=\linewidth]
        {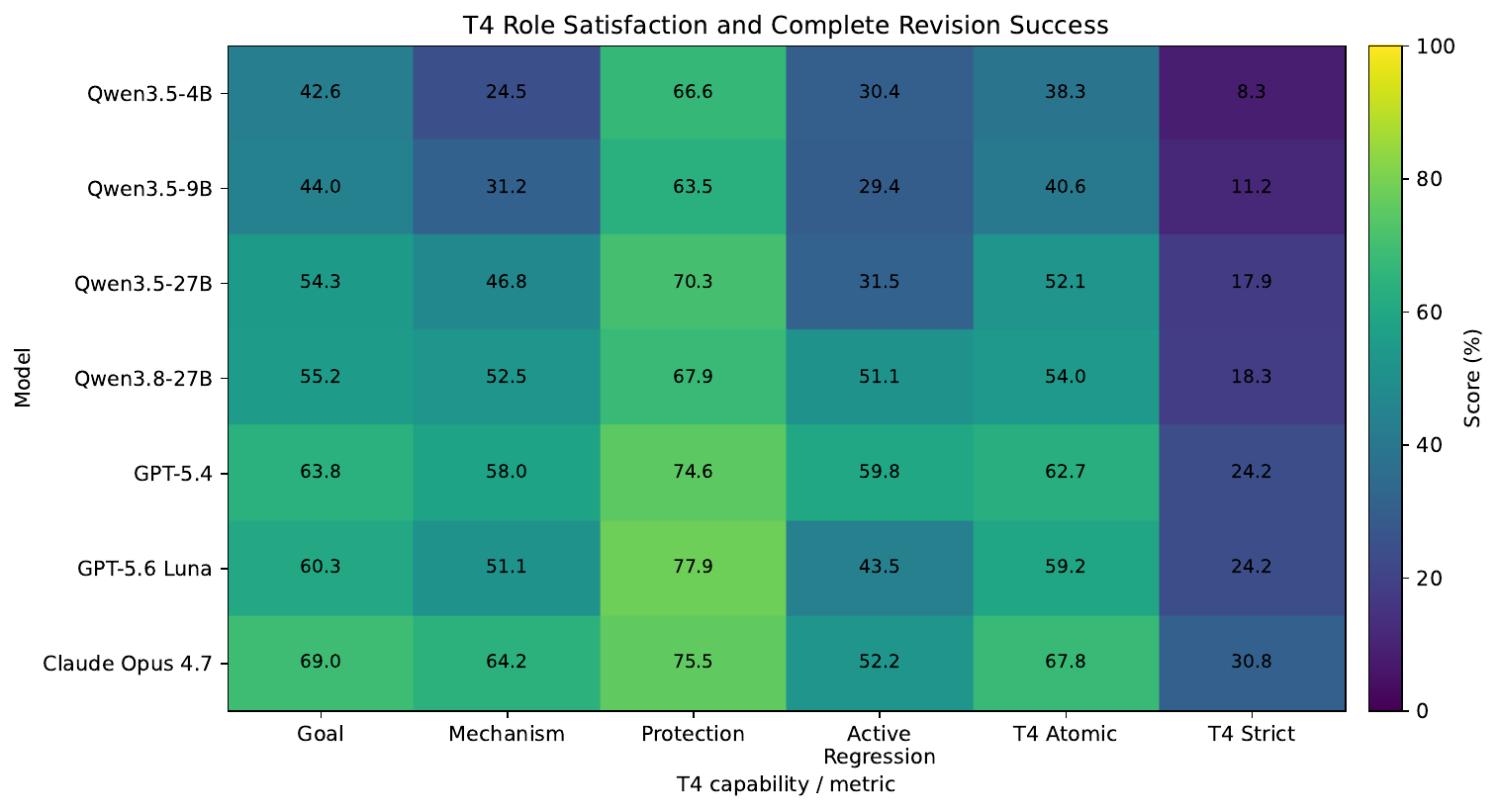}
         \vspace{-8mm}
        \caption{
        {Fine-grained T4 capability analysis.}
        }
        \label{fig:t4_role}
           \vspace{-12mm}
    \end{minipage}
    \hfill
    \begin{minipage}[t]{0.49\linewidth}
        \centering
        \includegraphics[width=\linewidth]
        {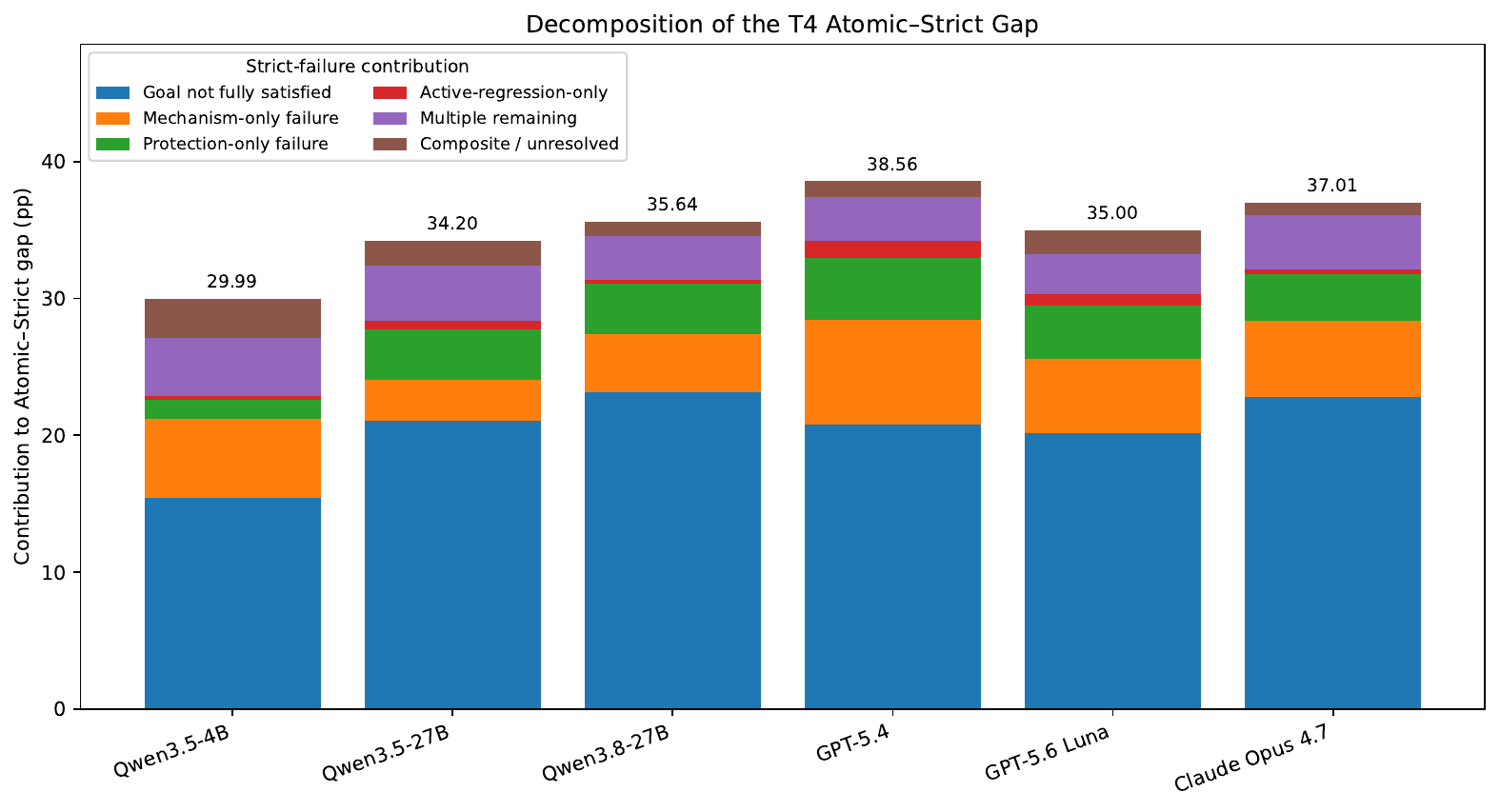}
        \vspace{-8mm}
        \caption{
        {The T4 Atomic--Strict gap.}
        }
        \vspace{-12mm}
        \label{fig:t4_gap}
    \end{minipage}
\vspace{-3mm}
\end{figure}

\textbf{Scaling improves revision capabilities unevenly.}
Within the Qwen3.5 family (Figure~\ref{fig:t4_role}), Mechanism
increases from 24.54\% at 4B to 31.19\% at 9B and 46.79\% at 27B,
while Goal increases from 42.65\% to 43.95\% and 54.32\%.
The official T4 Atomic score similarly rises from
38.33\% to 40.60\% and 52.12\%.
In contrast, Protection is non-monotonic
(66.56\%, 63.52\%, and 70.34\%),
and active Regression checking remains nearly unchanged
(30.43\%, 29.35\%, and 31.52\%).
Strict Success increases from 8.33\% to 11.25\% and 17.92\%,
but remains substantially below the local role scores.
The same heterogeneity appears among stronger API models.
Claude Opus 4.7 achieves the strongest Goal and Mechanism scores
(69.02\% and 64.22\%),
GPT-5.4 performs best on active Regression checking (59.78\%),
and GPT-5.6 Luna obtains the highest Protection score (77.94\%).

\textbf{Partial correctness does not imply complete revision validity.}
Even the strongest models exhibit a large difference between Atomic
and Strict performance:
Claude Opus 4.7 obtains 67.84\% Atomic but only 30.83\% Strict,
GPT-5.4 obtains 62.73\% versus 24.17\%,
and GPT-5.6 Luna obtains 59.16\% versus 24.17\%;
the gap is as large for open models (35.64 points for Qwen3.8-27B).
Figure~\ref{fig:t4_gap} decomposes this Atomic--Strict difference into
mutually exclusive strict-failure groups.
The decomposition represents partial Atomic credit retained by answers
that nevertheless fail the strict acceptance criterion.
Although a large fraction arises from responses that still fail at
least one Goal requirement, the remaining gap also includes responses
whose Goal is already satisfied but whose Mechanism, Protection,
Regression, or multiple residual conditions remain incomplete.
This distinction is confirmed by conditioning on responses that satisfy
all available Goal criteria.
For Claude, among 71 such responses with independently scored Protection
requirements, 25 still fail to satisfy all protection conditions;
among 114 with Mechanism requirements, 38 remain incomplete.
Detailed conditional failure statistics are provided in
Appendix~\ref{app:t4_goal_conditioned}.

\textbf{Explicit regression checking remains weak.}
The analysis further distinguishes the absence of a flagged critical
error from explicit regression checking.
Negative critical-error criteria are rarely triggered, whereas positive
criteria that explicitly require models to inspect residual or newly
introduced risks achieve substantially lower scores.
On the 46 items with such positive criteria, active Regression checking
reaches 59.78\% for GPT-5.4, 52.17\% for Claude Opus 4.7, and 51.09\%
for Qwen3.8-27B.
The coverage of the diagnostic roles is not uniform.
Among the 240 T4 items, 237 contain independently observable Goal
criteria, 218 contain Mechanism criteria, 159 contain Protection
criteria, and 46 contain positive active-Regression criteria.
Thirty composite criteria from 18 items cannot be reliably decomposed
into a unique role and are not assigned to a role.
For this reason, each role is reported using its own eligible-item
denominator.
Appendix~\ref{app:t4_common_scope} reports a common-scope sensitivity analysis on the subset where all major roles are observable.

\vspace{-3mm}

\subsection{Can SFT Narrow the Engineering Reasoning Gap?}
\vspace{-2mm}
\label{sec:industry_sft}
\begin{wraptable}{r}{0.5\textwidth}
\vspace{-8mm}
\centering
\caption{SFT results on EngIntervene.}
\label{tab:industry_sft}

\setlength{\tabcolsep}{4.5pt}

\resizebox{\linewidth}{!}{
\begin{tabular}{llrrrrrr}
\toprule
Model
& Condition
& T1
& T2
& T3
& T4 Atomic
& T4 Strict
& Macro Avg. \\
\midrule

\multirow{2}{*}{Qwen3.5-4B}
& Base
& 44.92 & 29.18 & 32.20 & 43.22 & 9.38 & 37.38 \\
& SFT
& \textbf{49.83} & 14.15 & 19.13 & 31.59 & 4.17 & 28.68 \\
\midrule

\multirow{2}{*}{InternVL3.5-8B}
& Base
& 38.55 & 22.35 & 22.05 & 40.41 & 8.33 & 30.84 \\
& SFT
& \textbf{43.69} & 13.05 & 16.73 & 34.17 & 5.21 & 26.91 \\
\midrule

\multirow{2}{*}{Qwen2.5-VL-7B}
& Base
& 42.46 & 18.07 & 16.31 & 32.98 & 4.17 & 27.45 \\
& SFT
& \textbf{45.92} & 12.70 & 14.77 & 28.02 & 4.17 & 25.35 \\

\bottomrule
\end{tabular}
}

\vspace{-3mm}
\end{wraptable}
To examine whether domain-specific supervision on EngIntervene can
directly improve engineering reasoning ability, we compare the original
base models with Industry-SFT variants for Qwen3.5-4B,
InternVL3.5-8B, and Qwen2.5-VL-7B in Table \ref{tab:industry_sft}.
Industry-SFT exhibits a highly consistent pattern of
{task differentiation} across all three models.
T1 improves in every case, indicating that domain-specific supervision
can improve adaptation to basic engineering objects, functions, and
states.
However, this improvement does not extend to the open-ended engineering
reasoning tasks. For all three models, Industry-SFT performs below the
corresponding Base model on {T2, T3, and T4 Atomic}.
Accordingly, the four-task macro average decreases from
{37.38\% to 28.68\%},
{30.84\% to 26.91\%}, and
{27.45\% to 25.35\%}, respectively.
T4 Strict also does not improve:
Qwen3.5-4B decreases from {9.38\% to 4.17\%},
InternVL3.5-8B from {8.33\% to 5.21\%},
while Qwen2.5-VL-7B remains unchanged at {4.17\%}.
These results show that, under the current LoRA-based Industry-SFT
setting, {learning domain-specific reference answers can improve
adaptation to basic tasks, but does not automatically translate into
stronger relation explanation, constraint diagnosis, or design-change
reasoning}.
In other words, the improvement on T1 is clearly decoupled from the
degradation on T2--T4, suggesting that basic domain adaptation and
higher-order engineering reasoning are not the same capability.
We therefore interpret the Industry-SFT results as evidence of
\textbf{negative transfer} under the current training setting.
The results indicate that the open-ended engineering reasoning gap in
EngIntervene cannot be easily eliminated through simple
benchmark-specific supervised imitation.
How to preserve and improve relational, constraint-aware, and
intervention reasoning during domain adaptation remains an open
question.

\vspace{-3mm}
\section{Related Work}
\vspace{-2mm}
\textbf{LLMs in Engineering Domains.}
Existing engineering benchmarks target problem solving, tool operation, or design generation: EngiBench \citep{zhou-etal-2026-engibench} covers multi-subfield engineering problem solving, FEABench \citep{mohammadzadeh2026fembenchstructuredscientificreasoning} has agents operate finite-element-analysis software, BikeBench \citep{regenwetter2025bikebenchbicycledesignbenchmark} evaluates bicycle design under multi-objective and multiphysics constraints, BuildArena \citep{xia2026buildarenaphysicsalignedinteractivebenchmark} has agents build structures in a physics simulator from natural-language goals, and AnalogCoder \citep{lai2025analogcoderprounifyinganalogcircuit} and Rule2DRC \citep{kim2026rule2drcbenchmarkingllmagents} evaluate code synthesis for analog circuits and DRC scripts.
\textsc{EngIntervene} evaluates the reasoning between understanding and generation: recovering the state of a given engineering system, diagnosing violated constraints, and determining how a local intervention changes the system while preserving required conditions.
\textbf{Multimodal Reasoning Benchmarks.}
Beyond visual recognition, MIRA \citep{XU2026109090} requires intermediate sketches or diagrams, DynaMath \citep{zou2025dynamathdynamicvisualbenchmark} tests robustness to visual and numerical variants of seed problems, ChartMuseum \citep{tang2026chartmuseumtestingvisualreasoning} targets reasoning over real-world charts, MMSI-Bench \citep{yang2026mmsibenchbenchmarkmultiimagespatial} evaluates multi-image spatial reasoning, SeePhys \citep{xiang2026seephys} and PhysBench \citep{chow2025physbenchbenchmarkingenhancingvisionlanguage} cover vision-essential physics and physical-world understanding, ComPABench \citep{li2025unveilingcompositionalabilitygap} examines compositional skill use under distribution shift, and MME-CoT \citep{jiang2025mmecotbenchmarkingchainofthoughtlarge} evaluates multi-step chain-of-thought reasoning across domains.
These benchmarks collect heterogeneous problems across broad domains; \textsc{EngIntervene} instead instantiates several reasoning levels, from visual grounding to design intervention, within one engineering setting, so that a model's use of a recovered state can be measured separately from its recovery.
\vspace{-4mm}
\section{Conclusion}
\vspace{-2mm}
We introduced \textsc{EngIntervene} to evaluate the missing middle between static multimodal engineering understanding and design generation: whether a model can use an understood engineering state to reason about relations, constraints, and design interventions. Its 3,229 questions across seven engineering domains instantiate this process in four levels, from state grounding to post-intervention validity. Our results show that recovering an engineering state does not guarantee its reliable use: better grounding does not consistently yield stronger diagnosis or intervention, visual evidence matters beyond recognition, benchmark-specific fine-tuning improves T1 but not T2--T4, and partial correctness on revision criteria often fails to produce a valid design change. Progress in multimodal engineering intelligence should therefore be measured by whether a recovered state supports constraint-aware decisions about how a system should change.

\section*{AI Use Statement}

Generative AI tools were used for language assistance, including translation and limited English polishing. All AI-assisted text was reviewed and revised by the authors. Large multimodal and language models were also used as experimental systems evaluated on \textsc{EngIntervene}, and model-based scoring was used for the open-ended T2--T4 tasks following the fixed evaluation protocol described in the paper.

\section*{Ethics Statement}

\textsc{EngIntervene} is constructed from traceable engineering materials, including examinations, textbooks, technical manuals, design guidelines, and other engineering reference sources. The benchmark does not intentionally collect personal or sensitive information, and its evaluation tasks focus on engineering systems rather than individuals. For non-English source materials, translation is performed during data normalization while preserving the original engineering meaning and source provenance.
Because some examples involve safety-relevant engineering domains such as aerospace, power systems, and building services, benchmark answers should not be treated as substitutes for professional engineering review or certification. For data release, we provide source provenance and will distribute benchmark materials in accordance with applicable source licenses and redistribution constraints.

\section*{Reproducibility Statement}

To support reproducibility, we will release the \textsc{EngIntervene} benchmark, evaluation code, scoring scripts, prompts, structured reference answers, and atomic rubrics through the anonymous project link provided with the submission. The release will also include the dataset splits, task and domain metadata, model inference settings, and scripts used to compute all reported metrics and ablations. For T2--T4, we will provide the fixed model-based evaluation protocol and scoring configuration used in the experiments. The anonymous project page will serve as a single entry point to the dataset and code repositories.

\bibliography{iclr2027_conference}

\begin{thebibliography}{24}
\providecommand{\natexlab}[1]{#1}
\providecommand{\url}[1]{\texttt{#1}}
\expandafter\ifx\csname urlstyle\endcsname\relax
  \providecommand{\doi}[1]{doi: #1}\else
  \providecommand{\doi}{doi: \begingroup \urlstyle{rm}\Url}\fi

\bibitem[Akbari et~al.(2026)Akbari, Gao, Zou, Yang, Duan, Torbunov, Wang, Ren, and Zhang]{akbari2026circuitsensehierarchicalmllmbenchmark}
Arman Akbari, Jian Gao, Yifei Zou, Mei Yang, Jinru Duan, Dmitrii Torbunov, Yanzhi Wang, Yihui Ren, and Xuan Zhang.
\newblock Circuitsense: A hierarchical mllm benchmark bridging visual comprehension and symbolic reasoning in engineering design process, 2026.
\newblock URL \url{https://arxiv.org/abs/2509.22339}.

\bibitem[Borgman(2000)]{electronicinfoinfra}
Christine~L. Borgman.
\newblock \emph{From Gutenberg to the Global Information Infrastructure: Access to Information in the Networked World}.
\newblock MIT Press, Cambridge, MA, 2000.
\newblock ISBN 978-0-262-02473-0.

\bibitem[Chow et~al.(2025)Chow, Mao, Li, Seita, Guizilini, and Wang]{chow2025physbenchbenchmarkingenhancingvisionlanguage}
Wei Chow, Jiageng Mao, Boyi Li, Daniel Seita, Vitor Guizilini, and Yue Wang.
\newblock Physbench: Benchmarking and enhancing vision-language models for physical world understanding, 2025.
\newblock URL \url{https://arxiv.org/abs/2501.16411}.

\bibitem[Doris et~al.(2024)Doris, Grandi, Tomich, Alam, Ataei, Cheong, and Ahmed]{doris2024designqamultimodalbenchmarkevaluating}
Anna~C. Doris, Daniele Grandi, Ryan Tomich, Md~Ferdous Alam, Mohammadmehdi Ataei, Hyunmin Cheong, and Faez Ahmed.
\newblock Designqa: A multimodal benchmark for evaluating large language models' understanding of engineering documentation, 2024.
\newblock URL \url{https://arxiv.org/abs/2404.07917}.

\bibitem[Groover(2021)]{manufacturing}
Mikell~P. Groover.
\newblock \emph{Fundamentals of Modern Manufacturing: Materials, Processes, and Systems}.
\newblock John Wiley \& Sons, 7 edition, 2021.
\newblock ISBN 978-1-119-70642-7.

\bibitem[Guo et~al.(2025)Guo, Li, Kong, Jiang, Zhao, Gong, Zhang, Li, Sang, Zhu, Jun, Huang, Liu, Xue, Kundu, Lim, Zhao, Granger, Younis, Keivan, Sabharwal, Sinha, Agarwal, Vandyck, Mai, Wang, Venkatesh, Barik, Yang, Yue, He, Wang, Xu, Chen, Wang, Xu, Shetty, Guo, Song, Jha, Liang, Yan, Zhang, Karnoor, Zhang, Pandya, Gong, Ganesh, Shi, Xu, Zhang, Ouyang, Qin, Rosenbaum, Snyder, Seiler, Dullerud, Zhang, Cheng, Hanumolu, Huang, Kulkarni, Namazifar, Zhang, and Hu]{guo2025engineeringagibenchmarkingengineering}
Xingang Guo, Yaxin Li, Xiangyi Kong, Yilan Jiang, Xiayu Zhao, Zhihua Gong, Yufan Zhang, Daixuan Li, Tianle Sang, Beixiao Zhu, Gregory Jun, Yingbing Huang, Yiqi Liu, Yuqi Xue, Rahul~Dev Kundu, Qi~Jian Lim, Yizhou Zhao, Luke~Alexander Granger, Mohamed~Badr Younis, Darioush Keivan, Nippun Sabharwal, Shreyanka Sinha, Prakhar Agarwal, Kojo Vandyck, Hanlin Mai, Zichen Wang, Aditya Venkatesh, Ayush Barik, Jiankun Yang, Chongying Yue, Jingjie He, Libin Wang, Licheng Xu, Hao Chen, Jinwen Wang, Liujun Xu, Rushabh Shetty, Ziheng Guo, Dahui Song, Manvi Jha, Weijie Liang, Weiman Yan, Bryan Zhang, Sahil~Bhandary Karnoor, Jialiang Zhang, Rutva Pandya, Xinyi Gong, Mithesh~Ballae Ganesh, Feize Shi, Ruiling Xu, Yifan Zhang, Yanfeng Ouyang, Lianhui Qin, Elyse Rosenbaum, Corey Snyder, Peter Seiler, Geir Dullerud, Xiaojia~Shelly Zhang, Zuofu Cheng, Pavan~Kumar Hanumolu, Jian Huang, Mayank Kulkarni, Mahdi Namazifar, Huan Zhang, and Bin Hu.
\newblock Toward engineering agi: Benchmarking the engineering design capabilities of llms, 2025.
\newblock URL \url{https://arxiv.org/abs/2509.16204}.

\bibitem[Hensen \& Lamberts(2011)Hensen and Lamberts]{buildingops}
Jan L.~M. Hensen and Roberto Lamberts (eds.).
\newblock \emph{Building Performance Simulation for Design and Operation}.
\newblock Spon Press, London, 2011.
\newblock ISBN 978-0-415-47414-6.

\bibitem[Jiang et~al.(2025)Jiang, Zhang, Guo, Li, Qi, Chen, Wang, Jin, Guo, Yan, Zhang, Fu, Gao, and Li]{jiang2025mmecotbenchmarkingchainofthoughtlarge}
Dongzhi Jiang, Renrui Zhang, Ziyu Guo, Yanwei Li, Yu~Qi, Xinyan Chen, Liuhui Wang, Jianhan Jin, Claire Guo, Shen Yan, Bo~Zhang, Chaoyou Fu, Peng Gao, and Hongsheng Li.
\newblock Mme-cot: Benchmarking chain-of-thought in large multimodal models for reasoning quality, robustness, and efficiency, 2025.
\newblock URL \url{https://arxiv.org/abs/2502.09621}.

\bibitem[Kim et~al.(2026)Kim, Byun, Hwang, Park, and Song]{kim2026rule2drcbenchmarkingllmagents}
Jinuk Kim, Junsoo Byun, Donghwi Hwang, Seong-Jin Park, and Hyun~Oh Song.
\newblock Rule2drc: Benchmarking llm agents for drc script synthesis with execution-guided test generation, 2026.
\newblock URL \url{https://arxiv.org/abs/2605.15669}.

\bibitem[Lai et~al.(2025)Lai, Poddar, Lee, Chen, Hu, Yu, Luo, and Pan]{lai2025analogcoderprounifyinganalogcircuit}
Yao Lai, Souradip Poddar, Sungyoung Lee, Guojin Chen, Mengkang Hu, Bei Yu, Ping Luo, and David~Z. Pan.
\newblock Analogcoder-pro: Unifying analog circuit generation and optimization via multi-modal llms, 2025.
\newblock URL \url{https://arxiv.org/abs/2508.02518}.

\bibitem[Li et~al.(2025)Li, Zhang, Rao, and Cheng]{li2025unveilingcompositionalabilitygap}
Tianle Li, Jihai Zhang, Yongming Rao, and Yu~Cheng.
\newblock Unveiling the compositional ability gap in vision-language reasoning model, 2025.
\newblock URL \url{https://arxiv.org/abs/2505.19406}.

\bibitem[Mohammadzadeh et~al.(2026)Mohammadzadeh, Hamdi, Shor, and Lejeune]{mohammadzadeh2026fembenchstructuredscientificreasoning}
Saeed Mohammadzadeh, Erfan Hamdi, Joel Shor, and Emma Lejeune.
\newblock Fem-bench: A structured scientific reasoning benchmark for evaluating code-generating llms, 2026.
\newblock URL \url{https://arxiv.org/abs/2512.20732}.

\bibitem[Regenwetter et~al.(2025)Regenwetter, Obaideh, Chiotti, Lykourentzou, and Ahmed]{regenwetter2025bikebenchbicycledesignbenchmark}
Lyle Regenwetter, Yazan~Abu Obaideh, Fabien Chiotti, Ioanna Lykourentzou, and Faez Ahmed.
\newblock Bikebench: A bicycle design benchmark for generative models with objectives and constraints, 2025.
\newblock URL \url{https://arxiv.org/abs/2508.00830}.

\bibitem[Schock et~al.(2012)Schock, Sims, Bull, Larsen, Likhachev, Nagano, Nilsson, Vuori, Yeager, Zhou, Zhang, and Weyant]{energy}
Robert~N. Schock, Ralph Sims, Stan Bull, Hans Larsen, Vladimir Likhachev, Koji Nagano, Hans Nilsson, Seppo Vuori, Kurt Yeager, Li~Zhou, Xiliang Zhang, and John Weyant.
\newblock Energy supply systems.
\newblock In Thomas~B. Johansson, Anand Patwardhan, Neboj{\v{s}}a Naki{\'c}enovi{\'c}, and Luis Gomez-Echeverri (eds.), \emph{Global Energy Assessment: Toward a Sustainable Future}, pp.\  1131--1172. Cambridge University Press, 2012.
\newblock \doi{10.1017/CBO9780511793677.021}.

\bibitem[Tang et~al.(2026)Tang, Kim, Zhao, Lake, Ding, Yin, Singhal, Wadhwa, Liu, Sprague, Namuduri, Hu, Rodriguez, Peng, and Durrett]{tang2026chartmuseumtestingvisualreasoning}
Liyan Tang, Grace Kim, Xinyu Zhao, Thom Lake, Wenxuan Ding, Fangcong Yin, Prasann Singhal, Manya Wadhwa, Zeyu~Leo Liu, Zayne Sprague, Ramya Namuduri, Bodun Hu, Juan~Diego Rodriguez, Puyuan Peng, and Greg Durrett.
\newblock Chartmuseum: Testing visual reasoning capabilities of large vision-language models, 2026.
\newblock URL \url{https://arxiv.org/abs/2505.13444}.

\bibitem[Vuchic(2007)]{transportation}
Vukan~R. Vuchic.
\newblock \emph{Urban Transit Systems and Technology}.
\newblock John Wiley \& Sons, 2007.
\newblock ISBN 978-0-471-75823-5.
\newblock \doi{10.1002/9780470168066}.

\bibitem[Xia et~al.(2026)Xia, Gao, Deng, Wei, Qian, Yu, and Wu]{xia2026buildarenaphysicsalignedinteractivebenchmark}
Tian Xia, Tianrun Gao, Wenhao Deng, Long Wei, Xiaowei Qian, Chenglei Yu, and Tailin Wu.
\newblock Buildarena: A physics-aligned interactive benchmark of llms for engineering construction, 2026.
\newblock URL \url{https://arxiv.org/abs/2510.16559}.

\bibitem[Xiang et~al.(2026)Xiang, Li, Zhang, Huang, Liu, Qu, He, Chen, Yuan, Han, Xu, Li, Sachan, and Liang]{xiang2026seephys}
Kun Xiang, Heng Li, Terry~Jingchen Zhang, Yinya Huang, Zirong Liu, Peixin Qu, Jixi He, Jiaqi Chen, Yu-Jie Yuan, Jianhua Han, Hang Xu, Hanhui Li, Mrinmaya Sachan, and Xiaodan Liang.
\newblock Seephys: Does seeing help thinking? {\textendash} benchmarking vision-based physics reasoning.
\newblock In \emph{The Thirty-ninth Annual Conference on Neural Information Processing Systems Datasets and Benchmarks Track}, 2026.
\newblock URL \url{https://openreview.net/forum?id=APNWmytTCS}.

\bibitem[XU et~al.(2026)XU, NIU, LI, Zhu, and Lu]{XU2026109090}
Qianhui XU, Ke~NIU, Jun LI, Shunzhe Zhu, and Yihang Lu.
\newblock Mira: Multi-scale invertible dual-attention redundancy-aware network for high-capacity video steganography.
\newblock \emph{Neural Networks}, 202:\penalty0 109090, 2026.
\newblock ISSN 0893-6080.
\newblock \doi{https://doi.org/10.1016/j.neunet.2026.109090}.
\newblock URL \url{https://www.sciencedirect.com/science/article/pii/S0893608026005502}.

\bibitem[Yang et~al.(2026)Yang, Xu, Xie, Yang, Li, Lin, Zhu, Chen, Duan, Yue, Lin, Wang, and Pang]{yang2026mmsibenchbenchmarkmultiimagespatial}
Sihan Yang, Runsen Xu, Yiman Xie, Sizhe Yang, Mo~Li, Jingli Lin, Chenming Zhu, Xiaochen Chen, Haodong Duan, Xiangyu Yue, Dahua Lin, Tai Wang, and Jiangmiao Pang.
\newblock Mmsi-bench: A benchmark for multi-image spatial intelligence, 2026.
\newblock URL \url{https://arxiv.org/abs/2505.23764}.

\bibitem[Yue et~al.(2024)Yue, Ni, Zhang, Zheng, Liu, Zhang, Stevens, Jiang, Ren, Sun, Wei, Yu, Yuan, Sun, Yin, Zheng, Yang, Liu, Huang, Sun, Su, and Chen]{yue2024mmmumassivemultidisciplinemultimodal}
Xiang Yue, Yuansheng Ni, Kai Zhang, Tianyu Zheng, Ruoqi Liu, Ge~Zhang, Samuel Stevens, Dongfu Jiang, Weiming Ren, Yuxuan Sun, Cong Wei, Botao Yu, Ruibin Yuan, Renliang Sun, Ming Yin, Boyuan Zheng, Zhenzhu Yang, Yibo Liu, Wenhao Huang, Huan Sun, Yu~Su, and Wenhu Chen.
\newblock Mmmu: A massive multi-discipline multimodal understanding and reasoning benchmark for expert agi, 2024.
\newblock URL \url{https://arxiv.org/abs/2311.16502}.

\bibitem[Zhou et~al.(2025)Zhou, Wang, Lin, and Preindl]{zhou2025software}
Liwei Zhou, Boya Wang, Zong-Han Lin, and Matthias Preindl.
\newblock Software-defined power electronics: Interconnection and coordination of power modules.
\newblock \emph{IEEE Journal of Emerging and Selected Topics in Power Electronics}, 2025.

\bibitem[Zhou et~al.(2026)Zhou, Wang, He, Zou, Wu, Cheng, Xie, Liu, Zhao, Xu, Gu, and Zhao]{zhou-etal-2026-engibench}
Xiyuan Zhou, Xinlei Wang, Yirui He, Ruixi Zou, Yang Wu, Yuheng Cheng, Yulu Xie, Wenxuan Liu, Huan Zhao, Yan Xu, Jinjin Gu, and Junhua Zhao.
\newblock {E}ngi{B}ench: A benchmark for evaluating large language models on engineering problem solving.
\newblock In Maria Liakata, Viviane~P. Moreira, Jiajun Zhang, and David Jurgens (eds.), \emph{Findings of the {A}ssociation for {C}omputational {L}inguistics: {ACL} 2026}, pp.\  36308--36334, San Diego, California, United States, July 2026. Association for Computational Linguistics.
\newblock ISBN 979-8-89176-395-1.
\newblock \doi{10.18653/v1/2026.findings-acl.1810}.
\newblock URL \url{https://aclanthology.org/2026.findings-acl.1810/}.

\bibitem[Zou et~al.(2025)Zou, Guo, Yang, Zhang, Hu, and Zhang]{zou2025dynamathdynamicvisualbenchmark}
Chengke Zou, Xingang Guo, Rui Yang, Junyu Zhang, Bin Hu, and Huan Zhang.
\newblock Dynamath: A dynamic visual benchmark for evaluating mathematical reasoning robustness of vision language models, 2025.
\newblock URL \url{https://arxiv.org/abs/2411.00836}.

\end{thebibliography}
\bibliographystyle{iclr2027_conference}

\clearpage

\section{Appendix}

\subsection{Qualitative Example}

\subsection{Task Definitions and Reference Answer Representations}
\label{app:task_definitions}

\subsubsection{T1: Multimodal Design Understanding}

A T1 evaluation unit consists of one or more images, a question, and candidate options. Correct answers are represented using stable option identifiers independent of displayed position. In addition to the correct option, the reference information may include the relevant visual object or local region and explanations provided by the source material. This task primarily covers component categories, functional roles, purposes of local features, and design intent.

\subsubsection{T2: Mechanistic and Relational Reasoning}

A T2 reference answer consists of required entities, critical relations, paths or sequences, and critical errors. Required entities identify components or regions that must appear in the answer; critical relations describe spatial, connectivity, motion, flow, load, thermal, electrical, or control relations among objects; paths or sequences represent multi-step processes; and critical errors identify statements involving reversed directions, missing paths, incorrect connections, or violations of basic engineering mechanisms.

\subsubsection{T3: Constraint-aware Design Diagnosis}

The diagnostic content of T3 can be summarized as
$
d=(l,i,c,e),$
where \(l\) denotes the problem location or object, \(i\) denotes the problem type, \(c\) denotes the violated constraint or design principle, and \(e\) denotes the potential consequence. When the question contains explicit numerical values, the reference representation additionally records the observed value and required value. A single question may involve multiple diagnostic elements, but the scoring scope covers only engineering judgments explicitly supported by the source material or question conditions.

\subsubsection{T4: Constraint-preserving Design Intervention Reasoning}

T4 reference answers are organized around design modifications and their consequences. Each sample describes at least the current problem or design objective and, where applicable, records the candidate or expected modification, target object of the modification, intended effect, mechanism, conditions that must remain satisfied, acceptable equivalent answers, and critical side effects that may invalidate the modification.
The four T4 subtypes correspond to different output requirements. Revision Selection primarily evaluates the selected candidate and its justification. Revision Verification and Regression Analysis requires determining the validity and completeness of a modification and whether it introduces new problems. Constrained Local Revision Proposal requires a concrete local modification satisfying the specified boundary conditions. Revision Prioritization and Trade-off requires ranking problems or modification directions according to explicit priorities.
Different expressions or implementations are considered equivalent when they produce the same engineering effect and simultaneously satisfy the target and protected conditions specified by the question.


\subsection{Complete Dataset Statistics}
\label{app:dataset_statistics}

\subsubsection{Complete Domain, Task, and T4 Subtype Distribution}
\label{app:t4_distribution}

Table~\ref{tab:domain_task_t4_subtypes} extends Fig.~\ref{tab:domain_task_distribution} by decomposing the 240 T4 questions into four design-intervention reasoning types across the seven engineering domains. Each T4 question is assigned to exactly one primary subtype; therefore, the four subtype counts sum to the T4 total for each domain.

The distribution of intervention types differs substantially across domains. All 30 T4 questions in Automotive System Integration are Revision Selection. Building Services and MEP Systems and Power and Energy Equipment contain no Revision Selection questions. Revision Verification and Regression Analysis and Constrained Local Revision Proposal each cover six domains. Revision Prioritization and Trade-off contains 25 questions distributed across Aerospace Integration and Reliability, Building Services and MEP Systems, Power and Energy Equipment, and Compact Electronics Packaging, with 5--7 questions per domain. The four T4 types therefore constitute a unified task taxonomy rather than four equally sized sub-benchmarks.

\begin{table*}[t]
\centering
\caption{\textbf{Complete domain-by-task and T4-subtype distribution.}}
\label{tab:domain_task_t4_subtypes}
\resizebox{\textwidth}{!}{
\begin{tabular}{lrrrrrrrrr}
\toprule
\textbf{Domain}
& \textbf{T1}
& \textbf{T2}
& \textbf{T3}
& \textbf{Selection}
& \textbf{Verification}
& \textbf{Proposal}
& \textbf{Prioritization}
& \textbf{T4 Total}
& \textbf{Overall} \\
\midrule
Mechanical Design, Manufacturing and Industrial Equipment
& 349 & 60 & 60 & 8 & 12 & 16 & 0 & 36 & 505 \\
Robotics and Mechatronics
& 300 & 60 & 60 & 13 & 22 & 1 & 0 & 36 & 456 \\
Automotive System Integration
& 300 & 60 & 60 & 30 & 0 & 0 & 0 & 30 & 450 \\
Aerospace Integration and Reliability
& 300 & 60 & 60 & 5 & 12 & 11 & 6 & 34 & 454 \\
Building Services and MEP Systems
& 300 & 60 & 60 & 0 & 23 & 11 & 6 & 40 & 460 \\
Power and Energy Equipment
& 300 & 60 & 60 & 0 & 12 & 15 & 7 & 34 & 454 \\
Compact Electronics Packaging
& 300 & 60 & 60 & 5 & 5 & 14 & 6 & 30 & 450 \\
\midrule
\textbf{Total}
& \textbf{2,149}
& \textbf{420}
& \textbf{420}
& \textbf{61}
& \textbf{86}
& \textbf{68}
& \textbf{25}
& \textbf{240}
& \textbf{3,229} \\
\bottomrule
\end{tabular}
}
\end{table*}

\subsubsection{Construction and Adaptation Statistics}
\label{app:construction_stats}

Table~\ref{tab:construction_stats} reports how questions are constructed across tasks. EngIntervene distinguishes normalized or constrained adaptations of source questions from source-supported constructed questions grounded in traceable engineering materials. A total of 2,850 questions correspond to normalized, translated, or cropped source questions, while 379 are source-supported constructed questions. The proportion of the latter ranges from 7.8\% to 11.0\% for T1--T3 and reaches 55.0\% for T4: higher-level design-modification questions therefore rely more heavily on controlled construction based on source evidence and explicit acceptance conditions.

\begin{table}[t]
\centering
\caption{\textbf{Construction and adaptation statistics.}
The Source-based column includes normalization, translation, cropping, and constrained adaptation.}
\label{tab:construction_stats}
\resizebox{\linewidth}{!}{
\begin{tabular}{lrrrr}
\toprule
\textbf{Task}
& \textbf{Source-based / Normalized / Constrained Adaptation}
& \textbf{Source-supported Constructed}
& \textbf{Total}
& \textbf{Constructed \%} \\
\midrule
T1 & 1,982 & 167 & 2,149 & 7.8\% \\
T2 & 386 & 34 & 420 & 8.1\% \\
T3 & 374 & 46 & 420 & 11.0\% \\
T4 & 108 & 132 & 240 & 55.0\% \\
\midrule
\textbf{Total}
& \textbf{2,850}
& \textbf{379}
& \textbf{3,229}
& \textbf{11.7\%} \\
\bottomrule
\end{tabular}
}
\end{table}

\subsubsection{Source Type Composition}
\label{app:source_types}

Table~\ref{tab:source_types} reports the type composition of the 357 source documents directly associated with benchmark questions. During source registration, 56 fine-grained source-type labels are used, such as ``national licensing examination with marking scheme'' and ``technician certification question bank with answers.'' Here, these labels are grouped into seven categories according to material type. Each source document belongs to one category, and each question is counted according to its primary source. The full mapping of the 56 labels to these categories is released with the dataset.

Examination materials with official answers or marking schemes form the largest category: national and official qualification or licensing examinations account for 59.7\% of source documents and 47.8\% of questions; trade, technician, and professional certification question banks account for 13.8\% of questions; and exam-preparation guides account for 2.9\%. University and college examinations, problem sets, and textbook exercises account for 10.5\% of questions; official training and trade manuals account for 12.4\%; and regulatory advisory circulars and manufacturer technical documents account for 5.0\%. In addition, one human-verified question--answer compilation contributes 249 questions (7.7\%), making it the largest single source by item count (Appendix~\ref{app:source_concentration}). Multiple source records from the same institution, examination system, or document series may be related; these dependencies are represented as source groups in Appendix~\ref{app:source_concentration}.

\begin{table*}[t]
\centering
\caption{\textbf{Source type composition.}
Source percentages use 357 source documents as the denominator, while item percentages use 3,229 questions. Displayed percentages may not sum exactly because of rounding.}
\label{tab:source_types}
\resizebox{\textwidth}{!}{
\begin{tabular}{lrrrrr}
\toprule
\textbf{Source Category}
& \textbf{Fine-grained Labels}
& \textbf{\# Source Records}
& \textbf{\% Records}
& \textbf{\# Items}
& \textbf{\% Items} \\
\midrule
Official licensing and qualification examinations with answer keys or mark schemes
& 21 & 213 & 59.7\% & 1,545 & 47.8\% \\
Trade, technician and professional certification question banks
& 9 & 61 & 17.1\% & 444 & 13.8\% \\
Certification study and exam-preparation guides
& 3 & 10 & 2.8\% & 93 & 2.9\% \\
University and college examinations, problem sets and textbook exercises with official solutions
& 10 & 58 & 16.2\% & 338 & 10.5\% \\
Official training manuals, trade manuals and skills-competition materials
& 8 & 10 & 2.8\% & 400 & 12.4\% \\
Regulatory advisory circulars and manufacturer technical documents
& 4 & 4 & 1.1\% & 160 & 5.0\% \\
Human-verified question--answer compilation
& 1 & 1 & 0.3\% & 249 & 7.7\% \\
\midrule
\textbf{Total}
& \textbf{56}
& \textbf{357}
& \textbf{100\%}
& \textbf{3,229}
& \textbf{100\%} \\
\bottomrule
\end{tabular}
}
\end{table*}

\subsubsection{Source Language Distribution}
\label{app:source_language}

Table~\ref{tab:source_language} reports the original-language distribution of the 357 source documents. The sources include English, German, Japanese, Chinese, Hindi--English mixed materials, and Traditional-Chinese--English mixed materials, with English sources contributing the largest number of questions. The table reports the language of the original source material; all benchmark inputs are in English, with non-English source materials translated into English during normalization.

\begin{table}[t]
\centering
\caption{\textbf{Source language statistics.}}
\label{tab:source_language}
\begin{tabular}{lrr}
\toprule
\textbf{Language}
& \textbf{\# Source Records}
& \textbf{\# Items from Sources} \\
\midrule
de & 19 & 55 \\
en & 216 & 2,051 \\
hi--en (mixed) & 43 & 301 \\
ja & 48 & 562 \\
zh & 15 & 172 \\
zh-Hant & 12 & 66 \\
zh-Hant/en & 4 & 22 \\
\midrule
\textbf{Total} & \textbf{357} & \textbf{3,229} \\
\bottomrule
\end{tabular}
\end{table}

\subsubsection{Visual Assets and Multi-image Inputs}
\label{app:visual_assets}

The benchmark questions reference 2,275 distinct images, producing 3,936 image references in total. Most questions use a single image; 483 are multi-image questions, accounting for 15.0\% of the benchmark. In addition, 644 images are shared by two or more questions.

\begin{table}[t]
\centering
\caption{\textbf{Visual asset statistics.}}
\label{tab:visual_assets}
\begin{tabular}{lr}
\toprule
\textbf{Metric} & \textbf{Value} \\
\midrule
Distinct images used by items & 2,275 \\
Total image references across items & 3,936 \\
Single-image questions & 2,746 \\
Two-image questions & 287 \\
\(\geq 3\)-image questions & 196 \\
Multi-image questions & 483 \\
Multi-image rate & 15.0\% \\
Images shared by \(>1\) item & 644 \\
\bottomrule
\end{tabular}
\end{table}

\subsubsection{Input Statistics by Task}
\label{app:input_statistics}

Table~\ref{tab:input_statistics} compares question length and the number of visual inputs across the four tasks. Question length is measured in characters. T1 questions are relatively short, with a median length of 35, whereas T4 has a median length of 114, consistent with its need to specify design objectives, candidate modifications, editable boundaries, or constraint conditions. All four tasks are dominated by single-image questions while retaining a subset of multi-image inputs.

\begin{table*}[t]
\centering
\caption{\textbf{Input statistics by task.}}
\label{tab:input_statistics}
\resizebox{\textwidth}{!}{
\begin{tabular}{lrrrrrrrr}
\toprule
\textbf{Task}
& \textbf{\# Items}
& \textbf{Question Length Median}
& \textbf{Q1--Q3}
& \textbf{Mean}
& \textbf{1 Image}
& \textbf{2 Images}
& \textbf{\(\geq 3\) Images}
& \textbf{MC Options Median} \\
\midrule
T1 & 2,149 & 35.0 & 22.0--54.0 & 47.7 & 1,799 & 186 & 164 & 4 \\
T2 & 420 & 47.5 & 34.0--86.8 & 84.0 & 371 & 35 & 14 & N/A \\
T3 & 420 & 45.0 & 35.0--71.0 & 90.6 & 364 & 38 & 18 & N/A \\
T4 & 240 & 114.0 & 63.8--136.0 & 134.9 & 212 & 28 & 0 & N/A \\
\bottomrule
\end{tabular}
}
\end{table*}

\subsubsection{Source Concentration and Independence}
\label{app:source_concentration}

The 3,229 questions are directly associated with 357 source documents, which are organized into 227 source groups according to relationships among the sources. The median number of questions per source document is 5, while the median number per source group is 4. The largest individual source contributes 249 questions, accounting for 7.7\% of the benchmark; the five and ten largest sources contribute 18.0\% and 24.7\%, respectively.

\begin{table}[t]
\centering
\caption{\textbf{Source concentration and independence.}}
\label{tab:source_concentration}
\begin{tabular}{lr}
\toprule
\textbf{Metric} & \textbf{Value} \\
\midrule
Source records & 357 \\
Connected source groups & 227 \\
Median items per source record & 5.0 \\
Mean items per source record & 9.0 \\
Max items from one source record & 249 \\
Median items per source group & 4.0 \\
Max source-group size & 300 \\
Top-1 source share & 7.7\% \\
Top-5 source share & 18.0\% \\
Top-10 source share & 24.7\% \\
\bottomrule
\end{tabular}
\end{table}

These statistics describe the source composition of the complete dataset. Experimental data splits are constructed at the source-group level and are described separately in the experimental setup.

\subsubsection{Dedicated Audit of T4 Design Interventions}
\label{app:t4_audit}

T4 reference answers describe not only the final modification but also the state changes and constraint consequences induced by the modification. Verifying only the modification itself is therefore insufficient for ensuring evaluation validity. We conduct an independent dedicated audit of all 240 T4 questions. Reviewers have relevant engineering backgrounds for the corresponding domains and did not participate in constructing the samples. Each question is examined along six dimensions:

\begin{enumerate}
    \item \textbf{Target definition}: whether the target problem, design requirement, or priority condition is clearly specified;
    \item \textbf{Intervention validity}: whether the reference modification is supported by the source material or question conditions and is technically executable;
    \item \textbf{Mechanism validity}: whether the mechanism described in the reference answer is consistent with the system structure and engineering principles;
    \item \textbf{Protected-constraint completeness}: whether all critical conditions that must remain satisfied after the modification are adequately recorded;
    \item \textbf{Regression-condition completeness}: whether important regression conditions that could invalidate the modification or create new problems elsewhere are sufficiently covered;
    \item \textbf{Scope and acceptance-condition validity}: for questions containing local editable regions, acceptance conditions, or priority requirements, whether these boundaries are clearly specified and consistent with the reference modification.
\end{enumerate}

Table~\ref{tab:t4_audit} reports the audit results before revision. The pass rates for target definition and reference-intervention validity are 99.2\% and 97.9\%, respectively, while mechanism validity reaches 96.7\%. Protected-constraint completeness and regression-condition completeness are slightly lower, at 95.4\% and 93.8\%. Overall, 218/240 questions (90.8\%) satisfy all applicable criteria. Failed items are primarily addressed through three forms of revision: adding missing protected constraints, clarifying the applicability boundary of a valid modification, and adding critical regression conditions that may cause the target modification to fail. All revised items are subsequently rechecked by reviewers.

\begin{table*}[t]
\centering
\caption{\textbf{Dedicated audit of T4 design interventions.}}
\label{tab:t4_audit}
\resizebox{\textwidth}{!}{
\begin{tabular}{lrrr}
\toprule
\textbf{Audit Dimension}
& \textbf{Passed / Total}
& \textbf{Pass Rate}
& \textbf{95\% CI} \\
\midrule
Target problem is clearly defined
& 238 / 240 & 99.2\% & 97.0--99.8 \\
Reference intervention is technically valid and supported
& 235 / 240 & 97.9\% & 95.2--99.1 \\
Intervention mechanism is technically valid
& 232 / 240 & 96.7\% & 93.6--98.3 \\
Protected constraints are complete
& 229 / 240 & 95.4\% & 92.0--97.4 \\
Regression conditions are sufficiently covered
& 225 / 240 & 93.8\% & 89.9--96.2 \\
Editable scope / acceptance conditions are well-defined
& 233 / 240 & 97.1\% & 94.1--98.6 \\
All applicable criteria satisfied
& 218 / 240 & 90.8\% & 86.5--93.9 \\
\bottomrule
\end{tabular}
}
\end{table*}

The distribution of failed items across the four subtypes exhibits distinct structural patterns. For Revision Selection, the main issue is whether engineering differences among candidate solutions are sufficiently clear. For Revision Verification and Regression Analysis, the main issue is incomplete coverage of secondary effects in the reference answer. For Constrained Local Revision Proposal, issues primarily involve insufficiently specified editable scopes or acceptance conditions. For Revision Prioritization and Trade-off, the main issue is whether the ranking rationale follows the hard constraints, risks, or priorities specified by the question.

\subsubsection{Shared Engineering Constraints across Domains}
\label{app:constraint_coverage}

Figure~\ref{fig:constraint_heatmap} in the main paper visualizes the coverage of shared engineering constraints across the seven engineering domains. This section reports the complete statistics used to generate the figure.

EngIntervene organizes cross-domain engineering constraints into eight high-level categories: Function / Performance, Safety / Reliability, Spatial / Clearance, Interface / Compatibility, Manufacturability / Assembly, Maintainability / Serviceability, Standards / Regulation, and Cost / Resource. Constraint annotations are applied only to the 660 T3 and T4 questions involving constraint reasoning and are determined from the question text and reference answers. Domain assignments follow Fig.~\ref{tab:domain_task_distribution}, with 90--100 T3--T4 questions per domain. A question may involve multiple constraint categories, so category counts cannot be summed to obtain the total number of questions in a domain. Across the 660 questions, there are 2,131 constraint labels, averaging 3.2 labels per question. Function / Performance is the most common category (533 questions, 80.8\%), followed by Safety / Reliability (389 questions, 58.9\%), while Cost / Resource is the least common (71 questions, 10.8\%).

\begin{table*}[t]
\centering
\caption{\textbf{Domain-by-shared-engineering-constraint coverage for T3--T4.}
Each cell reports the number of questions in the domain involving the corresponding constraint category. A question may belong to multiple categories, so row sums exceed the number of T3--T4 questions in the domain. The final column reports the average number of constraint categories per question.}
\label{tab:constraint_coverage}
\resizebox{\textwidth}{!}{
\begin{tabular}{lrrrrrrrrrr}
\toprule
\textbf{Domain}
& \textbf{T3--T4}
& \textbf{Function / Performance}
& \textbf{Safety / Reliability}
& \textbf{Spatial / Clearance}
& \textbf{Interface / Compatibility}
& \textbf{Manufacturability / Assembly}
& \textbf{Maintainability / Serviceability}
& \textbf{Standards / Regulation}
& \textbf{Cost / Resource}
& \textbf{Labels / Item} \\
\midrule
Mechanical Design, Manufacturing and Industrial Equipment
& 96 & 82 & 28 & 46 & 24 & 58 & 31 & 8 & 14 & 3.0 \\
Robotics and Mechatronics
& 96 & 78 & 32 & 41 & 29 & 22 & 24 & 6 & 9 & 2.5 \\
Automotive System Integration
& 90 & 74 & 48 & 44 & 43 & 27 & 39 & 12 & 8 & 3.3 \\
Aerospace Integration and Reliability
& 94 & 81 & 72 & 38 & 34 & 41 & 36 & 47 & 7 & 3.8 \\
Building Services and MEP Systems
& 100 & 63 & 79 & 52 & 22 & 26 & 44 & 39 & 6 & 3.3 \\
Power and Energy Equipment
& 94 & 76 & 84 & 27 & 37 & 19 & 28 & 24 & 12 & 3.3 \\
Compact Electronics Packaging
& 90 & 79 & 46 & 35 & 42 & 55 & 29 & 9 & 15 & 3.4 \\
\midrule
\textbf{Total}
& \textbf{660}
& \textbf{533}
& \textbf{389}
& \textbf{283}
& \textbf{231}
& \textbf{248}
& \textbf{231}
& \textbf{145}
& \textbf{71}
& \textbf{3.2} \\
\bottomrule
\end{tabular}
}
\end{table*}

\subsubsection{Visual Representation and Engineering Terminology}
\label{app:visual_representation}

{Visual Representation Types}

The visual inputs in EngIntervene are grouped into nine primary representation types: CAD and rendered views, engineering drawings, assembly drawings, mechanism and system schematics, piping and wiring diagrams, BIM/MEP layouts, before/after comparison images, real-equipment images, and mixed representations composed of multiple visual types. Annotation is performed at the question level, and a question may belong to multiple categories. When the visual input combines two or more types and no dominant type can be reliably assigned, such as a combination of an engineering drawing and a real-world photograph, it is categorized as Mixed. Of the 3,229 questions, 3,115 are labeled; the remaining 114 questions (3.5\%) contain visual inputs involving multiple types that cannot be reliably categorized and are excluded from this analysis.

Real-equipment images are the most common visual representation (1,010 questions, 31.3\%), followed by engineering drawings (783, 24.2\%), piping and wiring diagrams (497, 15.4\%), and mixed representations (472, 14.6\%). Before/after comparisons account for 158 questions (4.9\%), while CAD/rendered views, assembly drawings, and BIM/MEP layouts together account for 11.2\%.

Visual-representation distributions vary across both tasks and domains. T2 is dominated by piping and wiring diagrams (156/411, 38.0\%), consistent with its emphasis on relational and path reasoning. T1, T3, and T4 are primarily represented by real-equipment images and engineering drawings. At the domain level, Mechanical Design and Aerospace are dominated by engineering drawings, which account for 53.7\% and 49.8\% of labeled questions in the respective domains. Automotive System Integration is dominated by engineering drawings and piping/wiring diagrams. Building Services and MEP primarily contains real-equipment images and BIM/MEP layouts; 97 of the 100 BIM/MEP-layout questions come from this domain. Power and Energy is dominated by real-equipment images (68.4\%). Robotics and Mechatronics contains the largest number of CAD views, with 89 questions accounting for 48.6\% of all CAD questions.

\begin{table}[t]
\centering
\caption{\textbf{Visual representation distribution.}
Percentages use all 3,229 questions as the denominator. Labels are multi-label, so the sum of category counts (3,542) exceeds the number of labeled questions (3,115), corresponding to an average of 1.14 labels per question.}
\label{tab:visual_representation}
\resizebox{0.5\linewidth}{!}{
\begin{tabular}{lrr}
\toprule
\textbf{Visual Representation}
& \textbf{\# Items}
& \textbf{\% Items} \\
\midrule
CAD / Rendered View & 183 & 5.7\% \\
Engineering Drawing & 783 & 24.2\% \\
Assembly Drawing & 78 & 2.4\% \\
Mechanism / System Schematic & 261 & 8.1\% \\
Piping / Wiring Diagram & 497 & 15.4\% \\
BIM / MEP Layout & 100 & 3.1\% \\
Before / After & 158 & 4.9\% \\
Real Equipment Image & 1,010 & 31.3\% \\
Mixed (multiple types) & 472 & 14.6\% \\
\bottomrule
\end{tabular}
}
\end{table}

\begin{table*}[t]
\centering
\caption{\textbf{Visual representation by task.}
Labels are multi-label and category counts may overlap. Before/after comparisons and mixed representations are combined in the final column.}
\label{tab:visual_by_task}
\resizebox{\textwidth}{!}{
\begin{tabular}{lrrrrrrrrrr}
\toprule
\textbf{Task}
& \textbf{\# Items}
& \textbf{\# Labeled}
& \textbf{CAD}
& \textbf{Drawing}
& \textbf{Assembly}
& \textbf{Schematic}
& \textbf{Piping / Wiring}
& \textbf{BIM / MEP}
& \textbf{Photo}
& \textbf{Before / After + Mixed} \\
\midrule
T1 & 2,149 & 2,081 & 103 & 562 & 51 & 136 & 200 & 66 & 713 & 442 \\
T2 & 420 & 411 & 24 & 84 & 15 & 59 & 156 & 20 & 80 & 86 \\
T3 & 420 & 394 & 33 & 73 & 3 & 44 & 105 & 7 & 128 & 66 \\
T4 & 240 & 229 & 23 & 64 & 9 & 22 & 36 & 7 & 89 & 36 \\
\midrule
\textbf{Total}
& \textbf{3,229}
& \textbf{3,115}
& \textbf{183}
& \textbf{783}
& \textbf{78}
& \textbf{261}
& \textbf{497}
& \textbf{100}
& \textbf{1,010}
& \textbf{630} \\
\bottomrule
\end{tabular}
}
\end{table*}

\begin{table*}[t]
\centering
\caption{\textbf{Visual representation by domain.}
Labels are multi-label and category counts may overlap. Before/after comparisons and mixed representations are combined in the final column.}
\label{tab:visual_by_domain}
\resizebox{\textwidth}{!}{
\begin{tabular}{lrrrrrrrrrr}
\toprule
\textbf{Domain}
& \textbf{\# Items}
& \textbf{\# Labeled}
& \textbf{CAD}
& \textbf{Drawing}
& \textbf{Assembly}
& \textbf{Schematic}
& \textbf{Piping / Wiring}
& \textbf{BIM / MEP}
& \textbf{Photo}
& \textbf{Before / After + Mixed} \\
\midrule
Mechanical Design, Manufacturing and Industrial Equipment
& 505 & 488 & 31 & 262 & 32 & 45 & 25 & 0 & 117 & 67 \\
Robotics and Mechatronics
& 456 & 436 & 89 & 37 & 6 & 86 & 52 & 2 & 133 & 123 \\
Automotive System Integration
& 450 & 446 & 6 & 154 & 12 & 53 & 143 & 0 & 28 & 105 \\
Aerospace Integration and Reliability
& 454 & 442 & 4 & 220 & 23 & 31 & 76 & 0 & 30 & 127 \\
Building Services and MEP Systems
& 460 & 446 & 27 & 38 & 2 & 10 & 70 & 97 & 214 & 51 \\
Power and Energy Equipment
& 454 & 408 & 1 & 6 & 0 & 20 & 61 & 1 & 279 & 59 \\
Compact Electronics Packaging
& 450 & 449 & 25 & 66 & 3 & 16 & 70 & 0 & 209 & 98 \\
\midrule
\textbf{Total}
& \textbf{3,229}
& \textbf{3,115}
& \textbf{183}
& \textbf{783}
& \textbf{78}
& \textbf{261}
& \textbf{497}
& \textbf{100}
& \textbf{1,010}
& \textbf{630} \\
\bottomrule
\end{tabular}
}
\end{table*}

\subsection{Detailed T4 Diagnostic Results}
\label{app:t4_diagnostics}

\subsubsection{Role-wise T4 Performance}
\label{app:t4_role}

Table~\ref{tab:t4_role_scores} reports the complete role-wise T4 results.
For each role, we first average the mapped criterion judgments within
each eligible item and then macro-average across eligible items.
This item-macro definition prevents questions containing more decomposed
rubric criteria from receiving greater weight.

The role assignments used in this analysis were manually reviewed.
Goal, Mechanism, Protection, and Active Regression are observable in
237, 218, 159, and 46 T4 items, respectively.
The Atomic and Strict metrics are computed over all 240 T4
items and should therefore not be interpreted as averages of the four
diagnostic role scores.

\begin{table}[t]
\centering
\caption{
\textbf{Role-wise T4 performance (\%).}
Role columns report item-macro satisfaction scores over their respective
eligible item sets.
Atomic and Strict are the T4 metrics over all 240 items.
}
\label{tab:t4_role_scores}
\resizebox{\linewidth}{!}{
\begin{tabular}{lrrrrrr}
\toprule
\textbf{Model}
& \textbf{Goal}
& \textbf{Mechanism}
& \textbf{Protection}
& \textbf{Active Reg.}
& \textbf{Atomic}
& \textbf{Strict} \\
\midrule
Qwen3.5-4B              & 42.65 & 24.54 & 66.56 & 30.43 & 38.33 &  8.33 \\
Qwen3.5-9B              & 43.95 & 31.19 & 63.52 & 29.35 & 40.60 & 11.25 \\
Qwen3.5-27B             & 54.32 & 46.79 & 70.34 & 31.52 & 52.12 & 17.92 \\
Qwen3.6-27B             & 51.55 & 40.83 & 69.08 & 34.78 & 49.04 & 13.33 \\
Qwen3.8-27B             & 55.17 & 52.52 & 67.92 & 51.09 & 53.98 & 18.33 \\
Qwen3-VL-2B-Thinking    & 33.23 & 19.95 & 55.82 &  7.61 & 29.63 &  6.25 \\
Qwen3-VL-8B-Thinking    & 45.39 & 35.55 & 67.82 & 23.91 & 42.70 & 11.25 \\
Qwen3-VL-8B-Instruct    & 31.82 & 19.27 & 39.94 & 17.39 & 27.51 &  6.25 \\
GPT-4o-mini             & 33.72 & 20.87 & 58.07 & 13.04 & 31.29 &  7.50 \\
GPT-5.4                 & 63.75 & 58.03 & 74.63 & 59.78 & 62.73 & 24.17 \\
GPT-5.6 Luna            & 60.34 & 51.15 & 77.94 & 43.48 & 59.16 & 24.17 \\
Claude Opus 4.7         & 69.02 & 64.22 & 75.52 & 52.17 & 67.84 & 30.83 \\
Grok 4.3                & 43.60 & 30.05 & 68.29 & 23.91 & 40.67 & 12.08 \\
\bottomrule
\end{tabular}
}
\end{table}

\subsubsection{Failures after Goal Satisfaction}
\label{app:t4_goal_conditioned}

Table~\ref{tab:t4_goal_conditioned} considers only responses for which
all available Goal criteria are satisfied and examines which remaining
requirements still prevent Strict Success.
This analysis separates failure to identify the intended modification
from failure to reason through its downstream engineering consequences.

A substantial fraction of goal-satisfied responses remain incomplete in
Mechanism, Protection, or Active Regression.
For example, Qwen3.5-27B fails complete Mechanism requirements on
29 of 84 eligible responses and Protection requirements on 19 of 55.
Claude Opus 4.7 fails the corresponding requirements on
38 of 114 and 25 of 71 responses, respectively.
Because eligibility differs across roles, these conditional rates are
descriptive rather than controlled causal comparisons.

\begin{table}[t]
\centering
\caption{
\textbf{Remaining failures after all available Goal criteria are
satisfied.}
Each failure rate uses goal-satisfied responses containing the
corresponding independently observable role as its denominator.
}
\label{tab:t4_goal_conditioned}
\resizebox{\linewidth}{!}{
\begin{tabular}{lrrrr}
\toprule
\textbf{Model}
& \textbf{Goal-sat.}
& \textbf{Mechanism fail}
& \textbf{Protection fail}
& \textbf{Active Reg. fail} \\
\midrule
Qwen3.5-4B
& 76  & 43/71 (60.56\%)  & 11/41 (26.83\%) & 19/25 (76.00\%) \\
Qwen3.5-9B
& 68  & 27/62 (43.55\%)  & 13/41 (31.71\%) & 18/22 (81.82\%) \\
Qwen3.5-27B
& 91  & 29/84 (34.52\%)  & 19/55 (34.55\%) & 19/27 (70.37\%) \\
Qwen3.6-27B
& 87  & 40/81 (49.38\%)  & 15/44 (34.09\%) & 12/23 (52.17\%) \\
Qwen3.8-27B
& 90  & 32/85 (37.65\%)  & 19/51 (37.25\%) & 13/27 (48.15\%) \\
Qwen3-VL-2B-Thinking
& 57  & 31/53 (58.49\%)  &  9/29 (31.03\%) & 15/16 (93.75\%) \\
Qwen3-VL-8B-Thinking
& 74  & 32/71 (45.07\%)  & 15/40 (37.50\%) & 16/23 (69.57\%) \\
Qwen3-VL-8B-Instruct
& 52  & 24/47 (51.06\%)  & 11/28 (39.29\%) & 17/21 (80.95\%) \\
GPT-4o-mini
& 52  & 26/49 (53.06\%)  &  9/31 (29.03\%) & 12/12 (100.00\%) \\
GPT-5.4
& 120 & 41/106 (38.68\%) & 22/65 (33.85\%) & 13/34 (38.24\%) \\
GPT-5.6 Luna
& 110 & 34/96 (35.42\%)  & 17/62 (27.42\%) & 14/28 (50.00\%) \\
Claude Opus 4.7
& 126 & 38/114 (33.33\%) & 25/71 (35.21\%) & 13/31 (41.94\%) \\
Grok 4.3
& 75  & 33/68 (48.53\%)  & 11/41 (26.83\%) & 21/25 (84.00\%) \\
\bottomrule
\end{tabular}
}
\end{table}

\subsubsection{Performance across T4 Intervention Types}
\label{app:t4_subtypes}

Table~\ref{tab:t4_subtypes} reports Atomic and Strict performance for
Revision Selection, Revision Verification and Regression Analysis,
Constrained Local Revision Proposal, and Revision Prioritization and
Trade-off.
Overall T4 improvement does not imply uniform improvement across
intervention operations.
For example, Qwen3.8-27B has a higher overall T4 score than
Qwen3.5-27B, yet its Selection Atomic score is lower
(35.93\% vs.\ 43.31\%), while it performs better on Verification and
Proposal.
Prioritization contains only five items and is therefore included only
for descriptive completeness.

\begin{table}[t]
\centering
\caption{
\textbf{Performance by T4 intervention subtype
(Atomic / Strict, \%).}
Selection, Verification, Proposal, and Prioritization.
}
\label{tab:t4_subtypes}
\resizebox{\linewidth}{!}{
\begin{tabular}{lrrrr}
\toprule
\textbf{Model}
& \textbf{Selection}
& \textbf{Verification}
& \textbf{Proposal}
& \textbf{Prioritization} \\
\midrule
Qwen3.5-4B
& 24.95 / 6.56 & 46.07 / 10.38 & 39.00 / 7.35 & 28.00 / 0.00 \\
Qwen3.5-9B
& 25.55 / 3.28 & 44.76 / 11.32 & 46.91 / 17.65 & 50.00 / 20.00 \\
Qwen3.5-27B
& 43.31 / 16.39 & 52.65 / 17.92 & 58.91 / 19.12 & 56.00 / 20.00 \\
Qwen3.6-27B
& 39.02 / 13.11 & 53.44 / 14.15 & 52.27 / 13.24 & 34.00 / 0.00 \\
Qwen3.8-27B
& 35.93 / 9.84 & 59.95 / 19.81 & 60.71 / 23.53 & 56.00 / 20.00 \\
Qwen3-VL-2B-Thinking
& 18.01 / 4.92 & 33.82 / 5.66 & 32.60 / 7.35 & 42.00 / 20.00 \\
Qwen3-VL-8B-Thinking
& 29.98 / 9.84 & 49.42 / 15.09 & 44.30 / 7.35 & 34.00 / 0.00 \\
Qwen3-VL-8B-Instruct
& 22.87 / 4.92 & 26.62 / 5.66 & 34.35 / 8.82 & 10.00 / 0.00 \\
GPT-4o-mini
& 23.85 / 4.92 & 36.56 / 6.60 & 29.40 / 10.29 & 36.00 / 20.00 \\
GPT-5.4
& 64.16 / 39.34 & 58.79 / 15.09 & 67.21 / 23.53 & 68.00 / 40.00 \\
GPT-5.6 Luna
& 64.57 / 36.07 & 53.83 / 16.98 & 63.45 / 26.47 & 48.00 / 0.00 \\
Claude Opus 4.7
& 71.38 / 45.90 & 65.59 / 24.53 & 67.29 / 25.00 & 80.00 / 60.00 \\
Grok 4.3
& 38.72 / 13.11 & 39.87 / 9.43 & 43.27 / 14.71 & 46.00 / 20.00 \\
\bottomrule
\end{tabular}
}
\end{table}

\subsubsection{Domain-level T4 Performance and Role Coverage}
\label{app:t4_domains}

Table~\ref{tab:t4_domain_strict} reports T4 Strict Success across the
seven engineering domains for representative models.
These differences should not be interpreted directly as intrinsic
domain difficulty because both intervention subtype composition and
role observability vary across domains.
For example, all 30 Automotive T4 items belong to Revision Selection,
while Aerospace, Building Services and MEP, and Compact Electronics
contain no independently scored positive Active Regression criteria.

\begin{table}[t]
\centering
\caption{
\textbf{T4 Strict Success by engineering domain (\%).}
The table is descriptive and should not be interpreted as a ranking of
intrinsic domain difficulty.
}
\label{tab:t4_domain_strict}
\resizebox{\linewidth}{!}{
\begin{tabular}{lrrrrr}
\toprule
\textbf{Domain}
& \textbf{Qwen3.5-27B}
& \textbf{Qwen3.8-27B}
& \textbf{GPT-5.4}
& \textbf{Luna}
& \textbf{Claude} \\
\midrule
Mechanical Design          &  8.33 & 11.11 & 22.22 & 19.44 & 22.22 \\
Robotics \& Mechatronics   & 33.33 & 36.11 & 47.22 & 47.22 & 52.78 \\
Automotive                 & 10.00 &  0.00 & 40.00 & 43.33 & 40.00 \\
Aerospace                  &  8.82 & 11.76 &  8.82 &  8.82 & 20.59 \\
Building Services \& MEP   & 30.00 & 25.00 & 15.00 & 20.00 & 20.00 \\
Power \& Energy            & 23.53 & 35.29 & 26.47 & 20.59 & 29.41 \\
Compact Electronics        &  6.67 &  3.33 & 10.00 & 10.00 & 33.33 \\
\bottomrule
\end{tabular}
}
\end{table}

Table~\ref{tab:t4_domain_coverage} reports the corresponding role
coverage.
The uneven coverage explains why domain-specific role scores cannot be
interpreted as measurements over identical task compositions.

\begin{table}[t]
\centering
\caption{
\textbf{Number of T4 items with independently observable role criteria
in each engineering domain.}
}
\label{tab:t4_domain_coverage}
\resizebox{\linewidth}{!}{
\begin{tabular}{lrrrr}
\toprule
\textbf{Domain}
& \textbf{T4}
& \textbf{Mechanism}
& \textbf{Protection}
& \textbf{Active Regression} \\
\midrule
Mechanical Design        & 36 & 30 & 26 & 12 \\
Robotics \& Mechatronics & 36 & 28 &  8 &  5 \\
Automotive               & 30 & 22 & 10 & 10 \\
Aerospace                & 34 & 34 & 34 &  0 \\
Building Services \& MEP & 40 & 40 & 40 &  0 \\
Power \& Energy          & 34 & 34 & 11 & 19 \\
Compact Electronics      & 30 & 30 & 30 &  0 \\
\midrule
\textbf{Total}
& \textbf{240}
& \textbf{218}
& \textbf{159}
& \textbf{46} \\
\bottomrule
\end{tabular}
}
\end{table}

\subsubsection{Common-Scope Sensitivity Check}
\label{app:t4_common_scope}

Because the four diagnostic roles are not independently observable in
the same number of T4 items, we additionally evaluate a common subset
of 134 questions for which Goal, Mechanism, Protection, and
regression-related requirements are simultaneously represented and no
unresolved composite role is present.
This subset is used only as a sensitivity analysis and does not replace
the  full-set T4 results.

As shown in Table~\ref{tab:t4_common_scope}, the main qualitative pattern
remains: stronger models achieve substantially higher Goal and Mechanism
scores, while Strict Success remains considerably lower than Atomic
performance.
Some model rankings change on this restricted subset, confirming that
role-specific averages should be interpreted diagnostically rather than
as an alternative benchmark leaderboard.

\begin{table}[t]
\centering
\caption{
\textbf{Common-scope sensitivity analysis on 134 T4 items.}
All reported models are evaluated on the same restricted item subset.
}
\label{tab:t4_common_scope}
\resizebox{\linewidth}{!}{
\begin{tabular}{lrrrrrr}
\toprule
\textbf{Model}
& \textbf{N}
& \textbf{Goal}
& \textbf{Mechanism}
& \textbf{Protection}
& \textbf{Atomic}
& \textbf{Strict} \\
\midrule
Qwen3.5-4B      & 134 & 28.36 & 23.88 & 67.79 & 34.01 &  6.72 \\
Qwen3.5-27B     & 134 & 44.40 & 49.25 & 71.52 & 50.22 & 14.93 \\
Qwen3.8-27B     & 134 & 46.64 & 58.96 & 68.66 & 53.42 & 12.69 \\
GPT-5.4         & 134 & 51.12 & 61.94 & 75.87 & 58.04 & 13.43 \\
Claude Opus 4.7 & 134 & 58.96 & 67.91 & 75.06 & 63.54 & 20.90 \\
\bottomrule
\end{tabular}
}
\end{table}

\subsection{Complete Dataset Statistics}
\label{app:dataset_statistics}

\subsubsection{Complete Domain, Task, and T4 Subtype Distribution}
\label{app:t4_distribution}

Table~\ref{tab:domain_task_t4_subtypes} extends Fig.~\ref{tab:domain_task_distribution} by decomposing the 240 T4 questions into four design-intervention reasoning types across the seven engineering domains. Each T4 question is assigned to exactly one primary subtype; therefore, the four subtype counts sum to the T4 total for each domain.

The distribution of intervention types differs substantially across domains. All 30 T4 questions in Automotive System Integration are Revision Selection. Building Services and MEP Systems and Power and Energy Equipment contain no Revision Selection questions. Revision Verification and Regression Analysis and Constrained Local Revision Proposal each cover six domains. Revision Prioritization and Trade-off contains 25 questions distributed across Aerospace Integration and Reliability, Building Services and MEP Systems, Power and Energy Equipment, and Compact Electronics Packaging, with 5--7 questions per domain. The four T4 types therefore constitute a unified task taxonomy rather than four equally sized sub-benchmarks.

\begin{table*}[t]
\centering
\caption{\textbf{Complete domain-by-task and T4-subtype distribution.}}
\label{tab:domain_task_t4_subtypes}
\resizebox{\textwidth}{!}{
\begin{tabular}{lrrrrrrrrr}
\toprule
\textbf{Domain}
& \textbf{T1}
& \textbf{T2}
& \textbf{T3}
& \textbf{Selection}
& \textbf{Verification}
& \textbf{Proposal}
& \textbf{Prioritization}
& \textbf{T4 Total}
& \textbf{Overall} \\
\midrule
Mechanical Design, Manufacturing and Industrial Equipment
& 349 & 60 & 60 & 8 & 12 & 16 & 0 & 36 & 505 \\
Robotics and Mechatronics
& 300 & 60 & 60 & 13 & 22 & 1 & 0 & 36 & 456 \\
Automotive System Integration
& 300 & 60 & 60 & 30 & 0 & 0 & 0 & 30 & 450 \\
Aerospace Integration and Reliability
& 300 & 60 & 60 & 5 & 12 & 11 & 6 & 34 & 454 \\
Building Services and MEP Systems
& 300 & 60 & 60 & 0 & 23 & 11 & 6 & 40 & 460 \\
Power and Energy Equipment
& 300 & 60 & 60 & 0 & 12 & 15 & 7 & 34 & 454 \\
Compact Electronics Packaging
& 300 & 60 & 60 & 5 & 5 & 14 & 6 & 30 & 450 \\
\midrule
\textbf{Total}
& \textbf{2,149}
& \textbf{420}
& \textbf{420}
& \textbf{61}
& \textbf{86}
& \textbf{68}
& \textbf{25}
& \textbf{240}
& \textbf{3,229} \\
\bottomrule
\end{tabular}
}
\end{table*}

\subsubsection{Construction and Adaptation Statistics}
\label{app:construction_stats}

Table~\ref{tab:construction_stats} reports how questions are constructed across tasks. EngIntervene distinguishes normalized or constrained adaptations of source questions from source-supported constructed questions grounded in traceable engineering materials. A total of 2,850 questions correspond to normalized, translated, or cropped source questions, while 379 are source-supported constructed questions. The proportion of the latter ranges from 7.8\% to 11.0\% for T1--T3 and reaches 55.0\% for T4: higher-level design-modification questions therefore rely more heavily on controlled construction based on source evidence and explicit acceptance conditions.

\begin{table}[t]
\centering
\caption{\textbf{Construction and adaptation statistics.}
The Source-based column includes normalization, translation, cropping, and constrained adaptation.}
\label{tab:construction_stats}
\resizebox{\linewidth}{!}{
\begin{tabular}{lrrrr}
\toprule
\textbf{Task}
& \textbf{Source-based / Normalized / Constrained Adaptation}
& \textbf{Source-supported Constructed}
& \textbf{Total}
& \textbf{Constructed \%} \\
\midrule
T1 & 1,982 & 167 & 2,149 & 7.8\% \\
T2 & 386 & 34 & 420 & 8.1\% \\
T3 & 374 & 46 & 420 & 11.0\% \\
T4 & 108 & 132 & 240 & 55.0\% \\
\midrule
\textbf{Total}
& \textbf{2,850}
& \textbf{379}
& \textbf{3,229}
& \textbf{11.7\%} \\
\bottomrule
\end{tabular}
}
\end{table}

\subsubsection{Source Type Composition}
\label{app:source_types}

Table~\ref{tab:source_types} reports the type composition of the 357 source documents directly associated with benchmark questions. During source registration, 56 fine-grained source-type labels are used, such as ``national licensing examination with marking scheme'' and ``technician certification question bank with answers.'' Here, these labels are grouped into seven categories according to material type. Each source document belongs to one category, and each question is counted according to its primary source. The full mapping of the 56 labels to these categories is released with the dataset.

Examination materials with official answers or marking schemes form the largest category: national and official qualification or licensing examinations account for 59.7\% of source documents and 47.8\% of questions; trade, technician, and professional certification question banks account for 13.8\% of questions; and exam-preparation guides account for 2.9\%. University and college examinations, problem sets, and textbook exercises account for 10.5\% of questions; official training and trade manuals account for 12.4\%; and regulatory advisory circulars and manufacturer technical documents account for 5.0\%. In addition, one human-verified question--answer compilation contributes 249 questions (7.7\%), making it the largest single source by item count (Appendix~\ref{app:source_concentration}). Multiple source records from the same institution, examination system, or document series may be related; these dependencies are represented as source groups in Appendix~\ref{app:source_concentration}.

\begin{table*}[t]
\centering
\caption{\textbf{Source type composition.}
Source percentages use 357 source documents as the denominator, while item percentages use 3,229 questions. Displayed percentages may not sum exactly because of rounding.}
\label{tab:source_types}
\resizebox{\textwidth}{!}{
\begin{tabular}{lrrrrr}
\toprule
\textbf{Source Category}
& \textbf{Fine-grained Labels}
& \textbf{\# Source Records}
& \textbf{\% Records}
& \textbf{\# Items}
& \textbf{\% Items} \\
\midrule
Official licensing and qualification examinations with answer keys or mark schemes
& 21 & 213 & 59.7\% & 1,545 & 47.8\% \\
Trade, technician and professional certification question banks
& 9 & 61 & 17.1\% & 444 & 13.8\% \\
Certification study and exam-preparation guides
& 3 & 10 & 2.8\% & 93 & 2.9\% \\
University and college examinations, problem sets and textbook exercises with official solutions
& 10 & 58 & 16.2\% & 338 & 10.5\% \\
Official training manuals, trade manuals and skills-competition materials
& 8 & 10 & 2.8\% & 400 & 12.4\% \\
Regulatory advisory circulars and manufacturer technical documents
& 4 & 4 & 1.1\% & 160 & 5.0\% \\
Human-verified question--answer compilation
& 1 & 1 & 0.3\% & 249 & 7.7\% \\
\midrule
\textbf{Total}
& \textbf{56}
& \textbf{357}
& \textbf{100\%}
& \textbf{3,229}
& \textbf{100\%} \\
\bottomrule
\end{tabular}
}
\end{table*}

\subsubsection{Source Language Distribution}
\label{app:source_language}

Table~\ref{tab:source_language} reports the original-language distribution of the 357 source documents. The sources include English, German, Japanese, Chinese, Hindi--English mixed materials, and Traditional-Chinese--English mixed materials, with English sources contributing the largest number of questions. The table reports the language of the original source material; all benchmark inputs are in English, with non-English source materials translated into English during normalization.

\begin{table}[t]
\centering
\caption{\textbf{Source language statistics.}}
\label{tab:source_language}
\begin{tabular}{lrr}
\toprule
\textbf{Language}
& \textbf{\# Source Records}
& \textbf{\# Items from Sources} \\
\midrule
de & 19 & 55 \\
en & 216 & 2,051 \\
hi--en (mixed) & 43 & 301 \\
ja & 48 & 562 \\
zh & 15 & 172 \\
zh-Hant & 12 & 66 \\
zh-Hant/en & 4 & 22 \\
\midrule
\textbf{Total} & \textbf{357} & \textbf{3,229} \\
\bottomrule
\end{tabular}
\end{table}

\subsubsection{Visual Assets and Multi-image Inputs}
\label{app:visual_assets}

The benchmark questions reference 2,275 distinct images, producing 3,936 image references in total. Most questions use a single image; 483 are multi-image questions, accounting for 15.0\% of the benchmark. In addition, 644 images are shared by two or more questions.

\begin{table}[t]
\centering
\caption{\textbf{Visual asset statistics.}}
\label{tab:visual_assets}
\begin{tabular}{lr}
\toprule
\textbf{Metric} & \textbf{Value} \\
\midrule
Distinct images used by items & 2,275 \\
Total image references across items & 3,936 \\
Single-image questions & 2,746 \\
Two-image questions & 287 \\
\(\geq 3\)-image questions & 196 \\
Multi-image questions & 483 \\
Multi-image rate & 15.0\% \\
Images shared by \(>1\) item & 644 \\
\bottomrule
\end{tabular}
\end{table}

\subsubsection{Input Statistics by Task}
\label{app:input_statistics}

Table~\ref{tab:input_statistics} compares question length and the number of visual inputs across the four tasks. Question length is measured in characters. T1 questions are relatively short, with a median length of 35, whereas T4 has a median length of 114, consistent with its need to specify design objectives, candidate modifications, editable boundaries, or constraint conditions. All four tasks are dominated by single-image questions while retaining a subset of multi-image inputs.

\begin{table*}[t]
\centering
\caption{\textbf{Input statistics by task.}}
\label{tab:input_statistics}
\resizebox{\textwidth}{!}{
\begin{tabular}{lrrrrrrrr}
\toprule
\textbf{Task}
& \textbf{\# Items}
& \textbf{Question Length Median}
& \textbf{Q1--Q3}
& \textbf{Mean}
& \textbf{1 Image}
& \textbf{2 Images}
& \textbf{\(\geq 3\) Images}
& \textbf{MC Options Median} \\
\midrule
T1 & 2,149 & 35.0 & 22.0--54.0 & 47.7 & 1,799 & 186 & 164 & 4 \\
T2 & 420 & 47.5 & 34.0--86.8 & 84.0 & 371 & 35 & 14 & N/A \\
T3 & 420 & 45.0 & 35.0--71.0 & 90.6 & 364 & 38 & 18 & N/A \\
T4 & 240 & 114.0 & 63.8--136.0 & 134.9 & 212 & 28 & 0 & N/A \\
\bottomrule
\end{tabular}
}
\end{table*}

\subsubsection{Source Concentration and Independence}
\label{app:source_concentration}

The 3,229 questions are directly associated with 357 source documents, which are organized into 227 source groups according to relationships among the sources. The median number of questions per source document is 5, while the median number per source group is 4. The largest individual source contributes 249 questions, accounting for 7.7\% of the benchmark; the five and ten largest sources contribute 18.0\% and 24.7\%, respectively.

\begin{table}[t]
\centering
\caption{\textbf{Source concentration and independence.}}
\label{tab:source_concentration}
\begin{tabular}{lr}
\toprule
\textbf{Metric} & \textbf{Value} \\
\midrule
Source records & 357 \\
Connected source groups & 227 \\
Median items per source record & 5.0 \\
Mean items per source record & 9.0 \\
Max items from one source record & 249 \\
Median items per source group & 4.0 \\
Max source-group size & 300 \\
Top-1 source share & 7.7\% \\
Top-5 source share & 18.0\% \\
Top-10 source share & 24.7\% \\
\bottomrule
\end{tabular}
\end{table}

These statistics describe the source composition of the complete dataset. Experimental data splits are constructed at the source-group level and are described separately in the experimental setup.

\subsubsection{Dedicated Audit of T4 Design Interventions}
\label{app:t4_audit}

T4 reference answers describe not only the final modification but also the state changes and constraint consequences induced by the modification. Verifying only the modification itself is therefore insufficient for ensuring evaluation validity. We conduct an independent dedicated audit of all 240 T4 questions. Reviewers have relevant engineering backgrounds for the corresponding domains and did not participate in constructing the samples. Each question is examined along six dimensions:

\begin{enumerate}
    \item \textbf{Target definition}: whether the target problem, design requirement, or priority condition is clearly specified;
    \item \textbf{Intervention validity}: whether the reference modification is supported by the source material or question conditions and is technically executable;
    \item \textbf{Mechanism validity}: whether the mechanism described in the reference answer is consistent with the system structure and engineering principles;
    \item \textbf{Protected-constraint completeness}: whether all critical conditions that must remain satisfied after the modification are adequately recorded;
    \item \textbf{Regression-condition completeness}: whether important regression conditions that could invalidate the modification or create new problems elsewhere are sufficiently covered;
    \item \textbf{Scope and acceptance-condition validity}: for questions containing local editable regions, acceptance conditions, or priority requirements, whether these boundaries are clearly specified and consistent with the reference modification.
\end{enumerate}

Table~\ref{tab:t4_audit} reports the audit results before revision. The pass rates for target definition and reference-intervention validity are 99.2\% and 97.9\%, respectively, while mechanism validity reaches 96.7\%. Protected-constraint completeness and regression-condition completeness are slightly lower, at 95.4\% and 93.8\%. Overall, 218/240 questions (90.8\%) satisfy all applicable criteria. Failed items are primarily addressed through three forms of revision: adding missing protected constraints, clarifying the applicability boundary of a valid modification, and adding critical regression conditions that may cause the target modification to fail. All revised items are subsequently rechecked by reviewers.

\begin{table*}[t]
\centering
\caption{\textbf{Dedicated audit of T4 design interventions.}}
\label{tab:t4_audit}
\resizebox{\textwidth}{!}{
\begin{tabular}{lrrr}
\toprule
\textbf{Audit Dimension}
& \textbf{Passed / Total}
& \textbf{Pass Rate}
& \textbf{95\% CI} \\
\midrule
Target problem is clearly defined
& 238 / 240 & 99.2\% & 97.0--99.8 \\
Reference intervention is technically valid and supported
& 235 / 240 & 97.9\% & 95.2--99.1 \\
Intervention mechanism is technically valid
& 232 / 240 & 96.7\% & 93.6--98.3 \\
Protected constraints are complete
& 229 / 240 & 95.4\% & 92.0--97.4 \\
Regression conditions are sufficiently covered
& 225 / 240 & 93.8\% & 89.9--96.2 \\
Editable scope / acceptance conditions are well-defined
& 233 / 240 & 97.1\% & 94.1--98.6 \\
All applicable criteria satisfied
& 218 / 240 & 90.8\% & 86.5--93.9 \\
\bottomrule
\end{tabular}
}
\end{table*}

The distribution of failed items across the four subtypes exhibits distinct structural patterns. For Revision Selection, the main issue is whether engineering differences among candidate solutions are sufficiently clear. For Revision Verification and Regression Analysis, the main issue is incomplete coverage of secondary effects in the reference answer. For Constrained Local Revision Proposal, issues primarily involve insufficiently specified editable scopes or acceptance conditions. For Revision Prioritization and Trade-off, the main issue is whether the ranking rationale follows the hard constraints, risks, or priorities specified by the question.

\subsubsection{Shared Engineering Constraints across Domains}
\label{app:constraint_coverage}

Figure~\ref{fig:constraint_heatmap} in the main paper visualizes the coverage of shared engineering constraints across the seven engineering domains. This section reports the complete statistics used to generate the figure.

EngIntervene organizes cross-domain engineering constraints into eight high-level categories: Function / Performance, Safety / Reliability, Spatial / Clearance, Interface / Compatibility, Manufacturability / Assembly, Maintainability / Serviceability, Standards / Regulation, and Cost / Resource. Constraint annotations are applied only to the 660 T3 and T4 questions involving constraint reasoning and are determined from the question text and reference answers. Domain assignments follow Fig.~\ref{tab:domain_task_distribution}, with 90--100 T3--T4 questions per domain. A question may involve multiple constraint categories, so category counts cannot be summed to obtain the total number of questions in a domain. Across the 660 questions, there are 2,131 constraint labels, averaging 3.2 labels per question. Function / Performance is the most common category (533 questions, 80.8\%), followed by Safety / Reliability (389 questions, 58.9\%), while Cost / Resource is the least common (71 questions, 10.8\%).

\begin{table*}[t]
\centering
\caption{\textbf{Domain-by-shared-engineering-constraint coverage for T3--T4.}
Each cell reports the number of questions in the domain involving the corresponding constraint category. A question may belong to multiple categories, so row sums exceed the number of T3--T4 questions in the domain. The final column reports the average number of constraint categories per question.}
\label{tab:constraint_coverage}
\resizebox{\textwidth}{!}{
\begin{tabular}{lrrrrrrrrrr}
\toprule
\textbf{Domain}
& \textbf{T3--T4}
& \textbf{Function / Performance}
& \textbf{Safety / Reliability}
& \textbf{Spatial / Clearance}
& \textbf{Interface / Compatibility}
& \textbf{Manufacturability / Assembly}
& \textbf{Maintainability / Serviceability}
& \textbf{Standards / Regulation}
& \textbf{Cost / Resource}
& \textbf{Labels / Item} \\
\midrule
Mechanical Design, Manufacturing and Industrial Equipment
& 96 & 82 & 28 & 46 & 24 & 58 & 31 & 8 & 14 & 3.0 \\
Robotics and Mechatronics
& 96 & 78 & 32 & 41 & 29 & 22 & 24 & 6 & 9 & 2.5 \\
Automotive System Integration
& 90 & 74 & 48 & 44 & 43 & 27 & 39 & 12 & 8 & 3.3 \\
Aerospace Integration and Reliability
& 94 & 81 & 72 & 38 & 34 & 41 & 36 & 47 & 7 & 3.8 \\
Building Services and MEP Systems
& 100 & 63 & 79 & 52 & 22 & 26 & 44 & 39 & 6 & 3.3 \\
Power and Energy Equipment
& 94 & 76 & 84 & 27 & 37 & 19 & 28 & 24 & 12 & 3.3 \\
Compact Electronics Packaging
& 90 & 79 & 46 & 35 & 42 & 55 & 29 & 9 & 15 & 3.4 \\
\midrule
\textbf{Total}
& \textbf{660}
& \textbf{533}
& \textbf{389}
& \textbf{283}
& \textbf{231}
& \textbf{248}
& \textbf{231}
& \textbf{145}
& \textbf{71}
& \textbf{3.2} \\
\bottomrule
\end{tabular}
}
\end{table*}

\subsubsection{Visual Representation and Engineering Terminology}
\label{app:visual_representation}

{Visual Representation Types}

The visual inputs in EngIntervene are grouped into nine primary representation types: CAD and rendered views, engineering drawings, assembly drawings, mechanism and system schematics, piping and wiring diagrams, BIM/MEP layouts, before/after comparison images, real-equipment images, and mixed representations composed of multiple visual types. Annotation is performed at the question level, and a question may belong to multiple categories. When the visual input combines two or more types and no dominant type can be reliably assigned, such as a combination of an engineering drawing and a real-world photograph, it is categorized as Mixed. Of the 3,229 questions, 3,115 are labeled; the remaining 114 questions (3.5\%) contain visual inputs involving multiple types that cannot be reliably categorized and are excluded from this analysis.

Real-equipment images are the most common visual representation (1,010 questions, 31.3\%), followed by engineering drawings (783, 24.2\%), piping and wiring diagrams (497, 15.4\%), and mixed representations (472, 14.6\%). Before/after comparisons account for 158 questions (4.9\%), while CAD/rendered views, assembly drawings, and BIM/MEP layouts together account for 11.2\%.

Visual-representation distributions vary across both tasks and domains. T2 is dominated by piping and wiring diagrams (156/411, 38.0\%), consistent with its emphasis on relational and path reasoning. T1, T3, and T4 are primarily represented by real-equipment images and engineering drawings. At the domain level, Mechanical Design and Aerospace are dominated by engineering drawings, which account for 53.7\% and 49.8\% of labeled questions in the respective domains. Automotive System Integration is dominated by engineering drawings and piping/wiring diagrams. Building Services and MEP primarily contains real-equipment images and BIM/MEP layouts; 97 of the 100 BIM/MEP-layout questions come from this domain. Power and Energy is dominated by real-equipment images (68.4\%). Robotics and Mechatronics contains the largest number of CAD views, with 89 questions accounting for 48.6\% of all CAD questions.

\begin{table}[t]
\centering
\caption{\textbf{Visual representation distribution.}
Percentages use all 3,229 questions as the denominator. Labels are multi-label, so the sum of category counts (3,542) exceeds the number of labeled questions (3,115), corresponding to an average of 1.14 labels per question.}
\label{tab:visual_representation}
\resizebox{\linewidth}{!}{
\begin{tabular}{lrr}
\toprule
\textbf{Visual Representation}
& \textbf{\# Items}
& \textbf{\% Items} \\
\midrule
CAD / Rendered View & 183 & 5.7\% \\
Engineering Drawing & 783 & 24.2\% \\
Assembly Drawing & 78 & 2.4\% \\
Mechanism / System Schematic & 261 & 8.1\% \\
Piping / Wiring Diagram & 497 & 15.4\% \\
BIM / MEP Layout & 100 & 3.1\% \\
Before / After & 158 & 4.9\% \\
Real Equipment Image & 1,010 & 31.3\% \\
Mixed (multiple types) & 472 & 14.6\% \\
\bottomrule
\end{tabular}
}
\end{table}

\begin{table*}[t]
\centering
\caption{\textbf{Visual representation by task.}
Labels are multi-label and category counts may overlap. Before/after comparisons and mixed representations are combined in the final column.}
\label{tab:visual_by_task}
\resizebox{\textwidth}{!}{
\begin{tabular}{lrrrrrrrrrr}
\toprule
\textbf{Task}
& \textbf{\# Items}
& \textbf{\# Labeled}
& \textbf{CAD}
& \textbf{Drawing}
& \textbf{Assembly}
& \textbf{Schematic}
& \textbf{Piping / Wiring}
& \textbf{BIM / MEP}
& \textbf{Photo}
& \textbf{Before / After + Mixed} \\
\midrule
T1 & 2,149 & 2,081 & 103 & 562 & 51 & 136 & 200 & 66 & 713 & 442 \\
T2 & 420 & 411 & 24 & 84 & 15 & 59 & 156 & 20 & 80 & 86 \\
T3 & 420 & 394 & 33 & 73 & 3 & 44 & 105 & 7 & 128 & 66 \\
T4 & 240 & 229 & 23 & 64 & 9 & 22 & 36 & 7 & 89 & 36 \\
\midrule
\textbf{Total}
& \textbf{3,229}
& \textbf{3,115}
& \textbf{183}
& \textbf{783}
& \textbf{78}
& \textbf{261}
& \textbf{497}
& \textbf{100}
& \textbf{1,010}
& \textbf{630} \\
\bottomrule
\end{tabular}
}
\end{table*}

\begin{table*}[t]
\centering
\caption{\textbf{Visual representation by domain.}
Labels are multi-label and category counts may overlap. Before/after comparisons and mixed representations are combined in the final column.}
\label{tab:visual_by_domain}
\resizebox{\textwidth}{!}{
\begin{tabular}{lrrrrrrrrrr}
\toprule
\textbf{Domain}
& \textbf{\# Items}
& \textbf{\# Labeled}
& \textbf{CAD}
& \textbf{Drawing}
& \textbf{Assembly}
& \textbf{Schematic}
& \textbf{Piping / Wiring}
& \textbf{BIM / MEP}
& \textbf{Photo}
& \textbf{Before / After + Mixed} \\
\midrule
Mechanical Design, Manufacturing and Industrial Equipment
& 505 & 488 & 31 & 262 & 32 & 45 & 25 & 0 & 117 & 67 \\
Robotics and Mechatronics
& 456 & 436 & 89 & 37 & 6 & 86 & 52 & 2 & 133 & 123 \\
Automotive System Integration
& 450 & 446 & 6 & 154 & 12 & 53 & 143 & 0 & 28 & 105 \\
Aerospace Integration and Reliability
& 454 & 442 & 4 & 220 & 23 & 31 & 76 & 0 & 30 & 127 \\
Building Services and MEP Systems
& 460 & 446 & 27 & 38 & 2 & 10 & 70 & 97 & 214 & 51 \\
Power and Energy Equipment
& 454 & 408 & 1 & 6 & 0 & 20 & 61 & 1 & 279 & 59 \\
Compact Electronics Packaging
& 450 & 449 & 25 & 66 & 3 & 16 & 70 & 0 & 209 & 98 \\
\midrule
\textbf{Total}
& \textbf{3,229}
& \textbf{3,115}
& \textbf{183}
& \textbf{783}
& \textbf{78}
& \textbf{261}
& \textbf{497}
& \textbf{100}
& \textbf{1,010}
& \textbf{630} \\
\bottomrule
\end{tabular}
}
\end{table*}

\end{document}